\documentclass{article} 
\usepackage[preprint]{colm2026_conference}

\usepackage{microtype}
\usepackage{hyperref}
\usepackage{url}
\usepackage{booktabs}

\usepackage{lineno}
\usepackage{hyperref}
\usepackage{url}
\usepackage{graphicx}
\usepackage{booktabs}
\usepackage{array}
\usepackage{subcaption}
\usepackage{enumitem}
\usepackage[table]{xcolor} 
\usepackage[most]{tcolorbox}
\usepackage{multirow}
\usepackage{float}
\usepackage{titlesec}
\usepackage{listings} 
\usepackage{placeins}
\usepackage{fvextra}

\usepackage{xcolor}

\definecolor{darkblue}{rgb}{0, 0, 0.5}
\hypersetup{colorlinks=true, citecolor=darkblue, linkcolor=darkblue, urlcolor=darkblue}

\title{How Do AI Agents Spend Your Money? Analyzing and Predicting Token Consumption in Agentic Coding Tasks}

\author{Longju Bai\textsuperscript{1} \quad
  Zhemin Huang\textsuperscript{3,6} \quad
  Xingyao Wang\textsuperscript{4} \quad
  Jiao Sun\textsuperscript{5} \quad
  Rada Mihalcea\textsuperscript{1} \\
  \textbf{Erik Brynjolfsson}\textsuperscript{3} \quad
  \textbf{Alex Pentland}\textsuperscript{3,7} \quad
  \textbf{Jiaxin Pei}\textsuperscript{2}\thanks{Correspondence to longju@umich.edu and jiaxinpei@utexas.edu} \\
  \textsuperscript{1}University of Michigan \quad
  \textsuperscript{2}UT Austin \quad
  \textsuperscript{3}Stanford University \quad
  \textsuperscript{4}All Hands AI  \\
  \textsuperscript{5}Google Deepmind \quad
  \textsuperscript{6}Microsoft AI \quad
  \textsuperscript{7}MIT \quad
}

\begin{document}

\ifcolmsubmission
\linenumbers
\fi
\maketitle

\begin{abstract}
The wide adoption of AI agents in complex human workflows is driving rapid growth in LLM token consumption.\footnote{We use ``token consumption'' and ``token usage'' interchangeably to refer to both input and output tokens used by LLM agents.} When agents are deployed on tasks that require a significant amount of tokens, three questions naturally arise: \textbf{(1) Where do AI agents spend the tokens? (2) Which models are more token-efficient?} and \textbf{(3) Can agents predict their token usage before task execution?}
In this paper, we present the first systematic study of token consumption patterns in agentic coding tasks. We analyze trajectories from \textbf{eight frontier LLMs} on SWE-bench Verified and evaluate models' ability to predict their own token costs \emph{before} task execution. We find that: \textbf{(1)} agentic tasks are uniquely expensive, consuming 1000$\times$ more tokens than code reasoning and code chat, with input tokens rather than output tokens driving the overall cost; \textbf{(2)} token usage is \textbf{highly variable} and inherently stochastic: runs on the same task can differ by up to \textbf{30$\times$} in total tokens, and higher token usage does \textbf{not} translate into higher accuracy; instead, accuracy often peaks at intermediate cost and saturates at higher costs; \textbf{(3)} models vary substantially in token efficiency: on the same tasks, Kimi-K2 and Claude-Sonnet-4.5, on average, consume over 1.5 million more tokens than GPT-5; \textbf{(4)} task difficulty rated by human experts only weakly aligns with actual token costs, revealing a fundamental gap between human-perceived complexity and the computational effort agents actually expend; and \textbf{(5)} frontier models \textbf{fail to accurately predict} their own token usage (with weak-to-moderate correlations, up to 0.39) and systematically underestimate real token costs. Our study offers new insights into the economics of AI agents and can inspire future research in this direction. Code and data are available on \href{https://longjubai.github.io/agent_token_consumption/}{the project website}.
 \vspace{-0.5cm}
\end{abstract}

\begin{figure}[H]
    \centering
    \hspace*{-0.4cm}
    \includegraphics[width=0.9\textwidth]{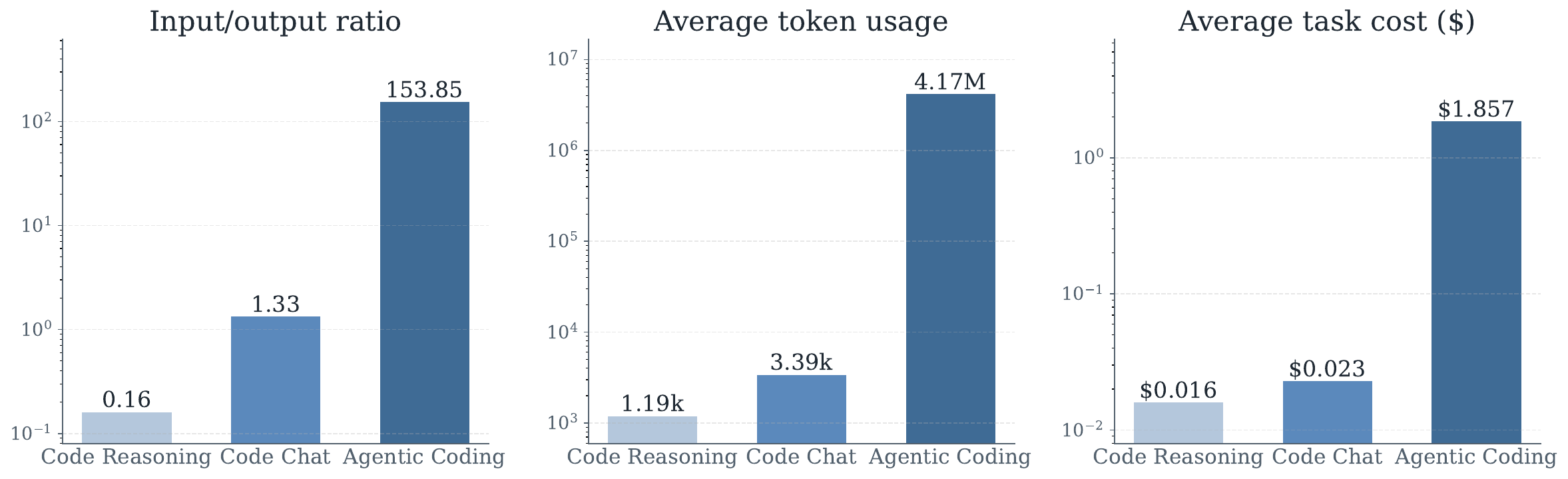}
    \caption{
    Agentic coding tasks cost significantly more tokens than \textit{Code Reasoning} (single-turn problem solving without tool interaction) and \textit{Code Chat} (multi-turn dialogue about a coding problem) tasks. Such a difference is largely driven by the increase of input tokens.}
    \label{fig:task_comparison}
\end{figure}

\section{Introduction}

Coding Agents are autonomous systems that can read repositories, reason about issues, call tools, and propose solutions with minimal human supervision~\citep{openai2025codex, liu2023repobenchbenchmarkingrepositorylevelcode, liu2023agentbenchevaluatingllmsagents,  jimenez2024swebench, wang2023voyager, wang2024executable}. While coding agents were originally developed mainly for coding tasks, due to their exceptional capabilities for using tools and working on long-horizon tasks, they have also been increasingly used in a wide range of tasks and domains beyond coding. 
Despite the wide adoption of coding agents and the productivity boost they bring, the prevailing pricing model for coding agents has been widely criticized for two reasons: (1) \textbf{lack of transparency}, users do not know the final cost until a task is finished; and (2) \textbf{no guarantee of completion}, users still need to pay for the token costs even if the task fails~\citep{kinde2024pricing}. These concerns converge on a central question: \textbf{Can we predict token consumption before a task is executed?} If we could estimate token usage up front, users would better understand potential costs and choose models accordingly; providers could also design clearer pricing tiers, enforce budget caps, and trigger early alerts for large bills.

In this paper, we present what is to our knowledge the first systematic study on AI Agent token consumption, complementing concurrent work on token distribution in multi-agent systems~\citep{salim2026tokenomics, wang2025agenttaxo}, and token pricing in reasoning models \citep{chen2026price}. To first understand the overall pattern of token usage in agentic coding tasks, we conduct an empirical study on trajectories generated by eight frontier LLMs using OpenHands agent~\citep{wang2025openhands} and SWE-bench-verified~\citep{jimenez2024swebench}. 
Our analysis reveals five key findings. First, {\bf agentic coding tasks are uniquely expensive}, consuming orders of magnitude more tokens than chat~\citep{codefeedback_sharegpt} and reasoning~\citep{pmlr-v235-gu24c} tasks (Figure~\ref{fig:task_comparison}). Strikingly, {\bf input tokens, not output tokens, dominate the overall cost} in agentic coding, even when token caching is enabled, consistent with recent analyses of token allocation across coding and reasoning tasks~\citep{wang2025agenttaxo, salim2026tokenomics}. Second, {\bf token usage is highly variable and inherently stochastic}: while more complex tasks tend to consume more tokens on average, usage varies substantially across runs, with some runs using up to $30\times$ more tokens than others on the same task. Third, {\bf more tokens do not translate into higher accuracy}: accuracy often peaks at intermediate cost and degrades at the highest cost levels, suggesting that excess token expenditure frequently reflects unproductive exploration rather than deeper reasoning. Fourth, {\bf models differ substantially in token efficiency}: on the same set of tasks, Kimi-K2 and Claude Sonnet-4.5 consume, on average, over 1.5 million more tokens than GPT-5. This gap holds even when restricting to the easy subset that all models solve successfully, showing that efficiency differences stem from model-specific behavior rather than intrinsic task difficulty. Finally, {\bf human-rated task difficulty only weakly aligns with actual token costs}, revealing a fundamental gap between human-perceived complexity and the computational effort agents actually expend. Together, these findings highlight both the heavy-tailed nature of token usage and the central role of context ingestion in agentic tasks.

Building on these observations, we further study agents' capabilities to predict their token costs before task execution.
We formalize the pre-execution agent token consumption predict tasks in which the agent is asked to predict input and output token usage given all its available tools and the coding environment. 
Rather than relying on static predictors or handcrafted features, the agent needs to act autonomously in the environment to produce cost estimates prior to execution.
We find that agents can capture coarse trends in token consumption, but achieve only \emph{weak-to-moderate} correlation with real usage across models. In general, \emph{output-token} usage is easier to predict than \emph{input-token} usage, reflecting the uncertainty introduced by context construction, retrieval, and tool-driven exploration. Additionally, \textbf{all the models systematically underestimate the actual token usage}, suggesting that token usage estimation is a very challenging task even for the most advanced models. Although accurate instance-level prediction remains challenging, self-prediction provides a useful coarse-grained signal of relative cost. This suggests that agent-driven estimation can potentially support early \emph{budget alerts} 
before launching expensive runs, improving cost transparency without overpromising precise token-level accuracy. 
Overall, our work makes the following contributions:
\begin{itemize}
\item We present the first large-scale empirical study of token consumption in agentic coding tasks, and open-source all agent trajectories from our experiments to support future research in this direction.
\item Our analysis reveals important insights into agent token consumption patterns that can inform future research and practice on agent pricing and model development.
\item We formulate the pre-execution agent token consumption prediction task and benchmark a range of frontier models, revealing a fundamental capability gap in estimating token usage before task execution.
\end{itemize}

Taken together, our empirical analysis and prediction study illuminate where tokens go in agentic coding and what can be anticipated before execution, providing concrete steps toward more predictable and user-aligned agent pricing.

\section{Data and Method}

We use \emph{OpenHands}~\citep{wang2025openhands} as our basic agent framework and collect agent trajectories on the SWE-Bench-Verified~\citep{jimenez2024swebench, chowdhury2024swebenchverified}, a benchmark of real-world GitHub issues paired with corresponding code repositories and tests.
Each problem is evaluated with four independent runs across a diverse set of LLMs, including Claude Sonnet-3.7, Sonnet-4, Sonnet-4.5, GPT-5, GPT-5.2, Qwen3-Coder-480B-A35B-Instruct, Kimi-K2, and Gemini-3-Pro.
We selected these models because they cover a diverse set of architectures, training paradigms, and deployment settings, while offering strong coding capabilities and reliable execution stability.

In this study, we focus specifically on token consumption throughout the end-to-end problem-solving process: given an initial task description, the LLM agent autonomously interacts with the environment to finish the task without any human intervention. For each problem instance, the agent proceeds in multiple rounds: in each round, the LLM generates a response based on the current prompt, followed by a tool call and execution. In particular, the full conversation history, including all previous prompts and completions, is carried forward unchanged into subsequent rounds.

To enable a more detailed analysis of LLM behavior and its corresponding token consumption during problem-solving, we extract a set of fine-grained metrics from the LLM completion history, such as per-type token cost, monetary cost, and action types. These metrics are obtained by parsing the structured JSON outputs of the agent and leveraging the usage information, which records all LLM interactions at each round. Together, the extracted metrics capture both the functional behavior of the agent, such as tool usage and file access patterns, as well as the underlying token-level dynamics. For all token-related metrics, we report values averaged over the four independent runs per problem.  The collected data include full execution trajectories, inference logs, intermediate outputs, evaluation results, and metadata, enabling a comprehensive analysis of agent behaviors and cost dynamics.

\section{Overall Agent Token Consumption Patterns}
In this section, we present key findings on agent token consumption in agentic coding tasks. We begin with a systematic comparison among agentic coding, code chat, and code reasoning tasks. Then we discuss the variances of token usage across tasks and runs, and whether higher token costs lead to more task completion. Finally, we discuss whether expert perception of task difficulty is aligned with actual agent token costs. 

\paragraph{Agentic tasks are uniquely expensive, and input tokens drive the cost of AI Agents}
We compare token usage across three coding-related tasks: code reasoning~\citep{pmlr-v235-gu24c}, coding chat~\citep{codefeedback_sharegpt}, and agentic coding~\citep{jimenez2024swebench, chowdhury2024swebenchverified}. Figure~\ref{fig:task_comparison} shows the average token usage, monetary costs, and input/output token ratio averaged across all the tasks. On average, agentic coding tasks consume \textbf{3500x} more tokens than a typical single-round reasoning task and \textbf{1200x} more tokens than a multi-round chatting task. Such a gap is primarily driven by the exponential growth of input tokens. Agentic workflows accumulate the information from different sources and the same context gets fed into the models repeatedly, resulting in a dramatically higher input/output ratio than the other two task types and significantly higher costs even with token caching.  Such a result reveals that agentic tasks are fundamentally different from other types of tasks and further motivates our study on agent token consumption.

\begin{figure*}[t]
\centering
\begin{subfigure}[t]{0.48\textwidth}
    \centering
    \includegraphics[width=\linewidth]{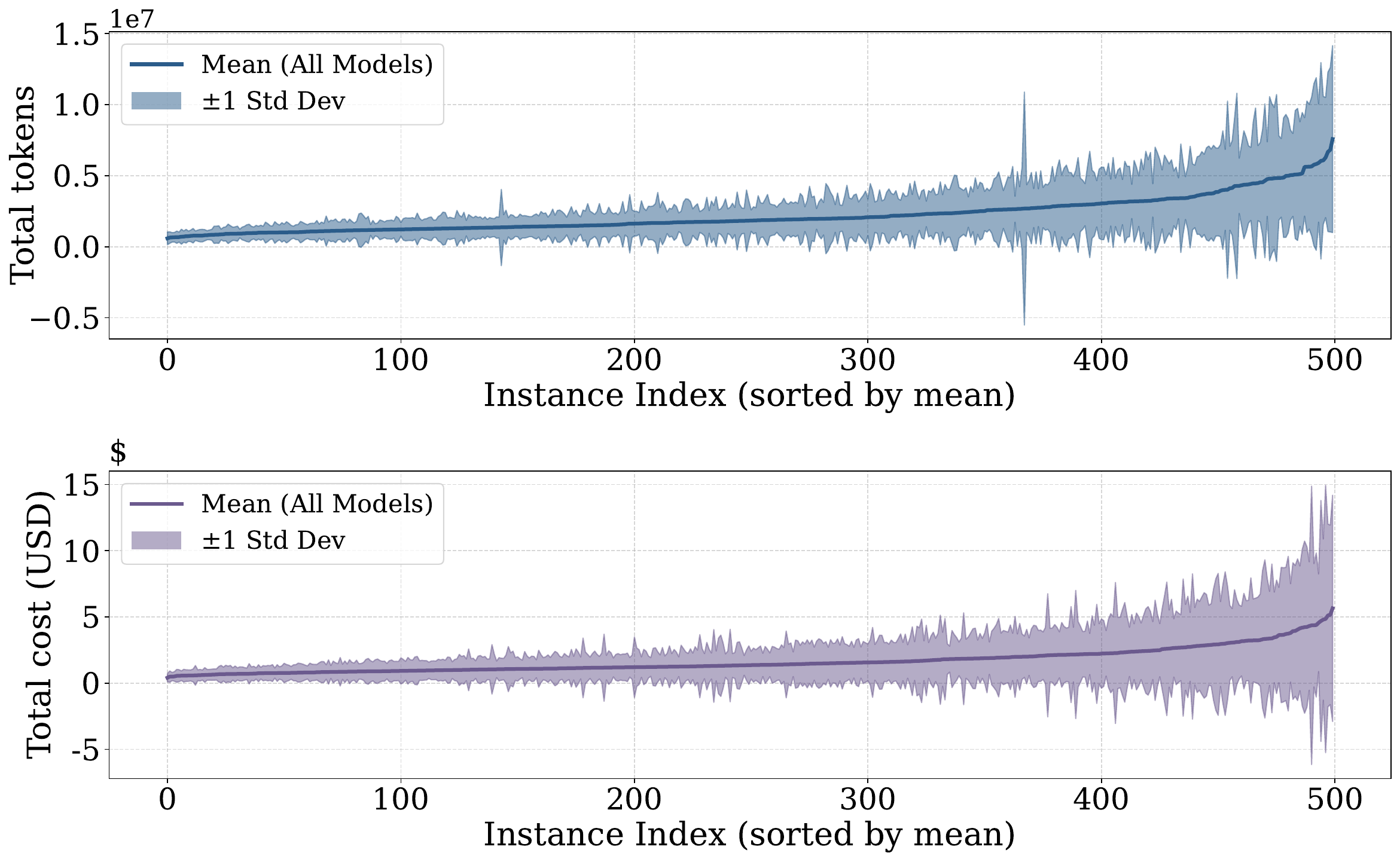}
    \subcaption{Across 500 problems, the most expensive instance costs $\sim$7M more tokens than the cheapest.}
    \label{fig:multi-model-variance}
\end{subfigure}
\hfill
\begin{subfigure}[t]{0.48\textwidth}
    \centering
    \includegraphics[width=\linewidth]{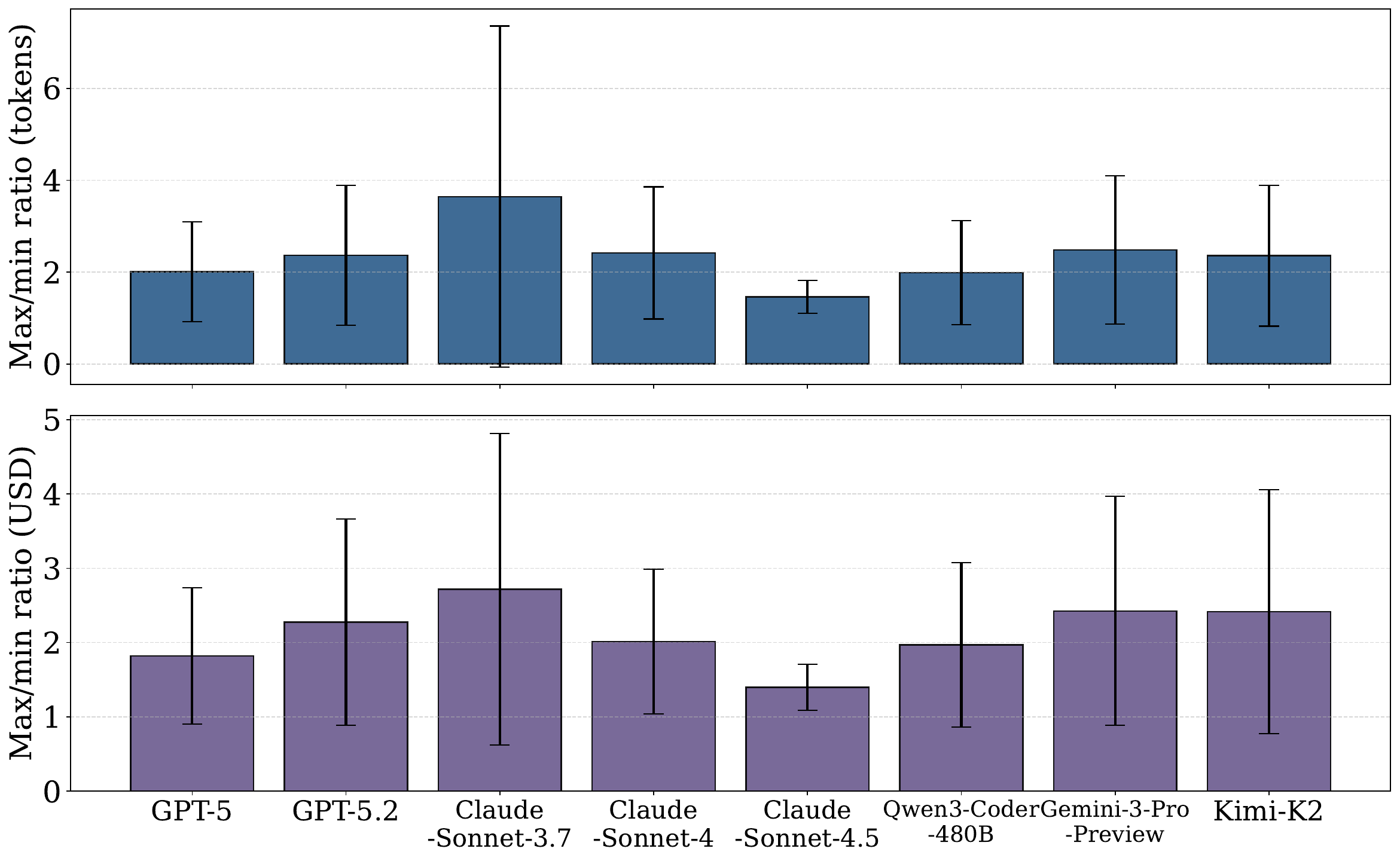}
    \subcaption{On the same problem, the most expensive run is $\sim$2$\times$ the cheapest across all models.}
    \label{fig:max-min-ratio}
\end{subfigure}
\caption{\textbf{Token cost is highly variable both across problems and across repeated runs of the same problem.}
(a) Per-instance mean $\pm$1\,SD across the four runs and eight models, with instances sorted by mean cost; the heavy right tail indicates that high-cost problems also exhibit the largest cross-run variance.
(b) Per-model max/min cost ratio, averaged across the 500 instances; error bars show $\pm$1\,SD across instances.
Together, these results suggest that token cost is highly variable, making upfront cost prediction fundamentally difficult.}
\label{fig:variance-combined}
\end{figure*}

\paragraph{Token usage is highly variable across problems and runs}
Are certain tasks costing more tokens, and do agents consume a similar amount of tokens when they are working on the same problem again? We analyze the variances of agent token usage across different problems (averaged over four independent runs) and across different runs of the same problem. 
Figure~\ref{fig:multi-model-variance} shows the aggregated agent token usage and cost across problems and runs.  We find that agents' token usage has large variances across different problems. The most expensive problem, on average, costs $\sim$7 million more tokens than the cheapest problem. Furthermore,  high-token-cost problems exhibit larger cross-run variance, indicating that agent behavior becomes increasingly unstable on more complex tasks. Figure~\ref{fig:max-min-ratio} shows the token usage difference for the most and least expensive runs for the same agent and problem. In general, the most expensive runs double the token and monetary cost of the least expensive runs, suggesting that the agent's token consumption has large variances even when working on exactly the same problem. Together, these results suggest that the token costs have large variances across problems and runs, making token usage prediction and agent pricing a fundamentally challenging task. 

\paragraph{More tokens do not lead to higher success rates}

Given the large variances of agent token consumption, one may wonder whether higher token usage leads to better performance. 
We first study this question at the problem level: whether tasks consuming more tokens have lower overall accuracy.
As shown in Figure~\ref{fig:multi-model-variance}, problems costing more input tokens have overall lower accuracy, and such a pattern is consistent across different models.  One intuitive explanation for this result is that more difficult tasks may naturally be more complicated, which further leads to higher token consumption. A similar trend is also observed for output tokens and we present the results in Appendix~\ref{sec:completion-tokens}.

When a coding agent is working on a specific task multiple times, the user may naturally expect the runs that cost more tokens to lead to higher accuracy. But do high-token-usage runs actually lead to higher accuracy? In our experiment, we run the same agent on a single problem four times. 
For each problem, we rank the four runs by token cost and group them into four categories: \emph{MinCost}, \emph{LowerCost}, \emph{UpperCost}, and \emph{MaxCost}. As shown in Figure~\ref{fig:max-min-ratio}, accuracy increases modestly from \emph{MinCost} to \emph{LowerCost}, but then saturates in higher-cost settings.
This non-monotonic trend is consistent with recent findings on \emph{inverse test-time scaling}~\citep{snell2024scalingllmtesttimecompute, wu2025inferencescalinglawsempirical, gema2025inverse, zeng2025revisiting, yang2025towards, aggarwal2025optimalthinkingbench}, which show that additional reasoning steps or longer chains of thought do not necessarily improve accuracy and may instead amplify distractors, spurious correlations, or inefficient reasoning cycles. 
In agentic settings, similar efficiency–performance trade-offs have been observed in long-horizon or ensemble-style systems, where increased computation does not reliably translate to improved task resolution~\citep{fan2025swe, wang2025efficient}. Our results provide further evidence along this line that \textbf{simply scaling token usage may not lead to higher execution performance. }

\begin{figure*}[t]
  \centering
  \begin{subfigure}{0.47\textwidth}
    \centering
    \includegraphics[width=\linewidth]{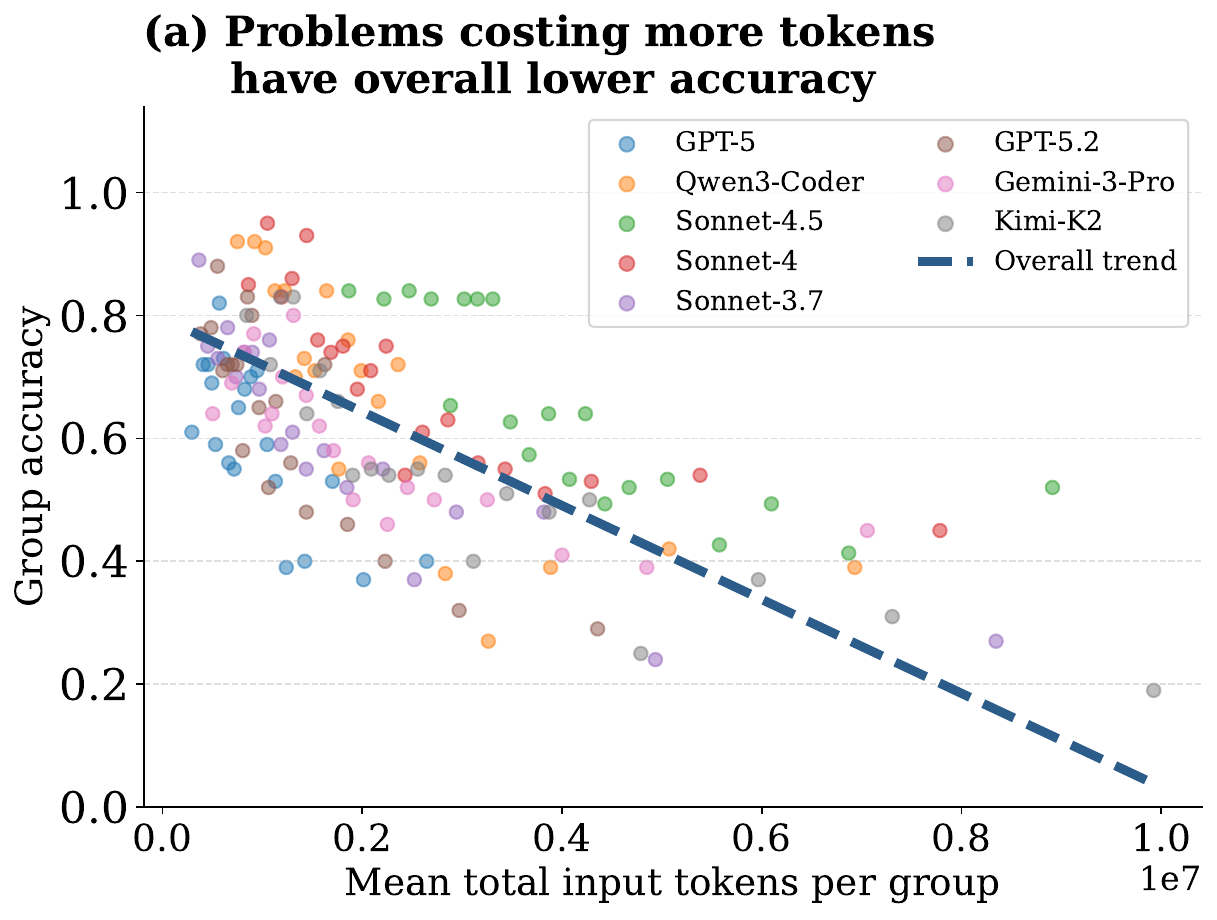}
    \label{fig:prompt-scatter}
  \end{subfigure}
  \hfill
  \begin{subfigure}{0.47\textwidth}
    \centering
    \includegraphics[width=\linewidth]{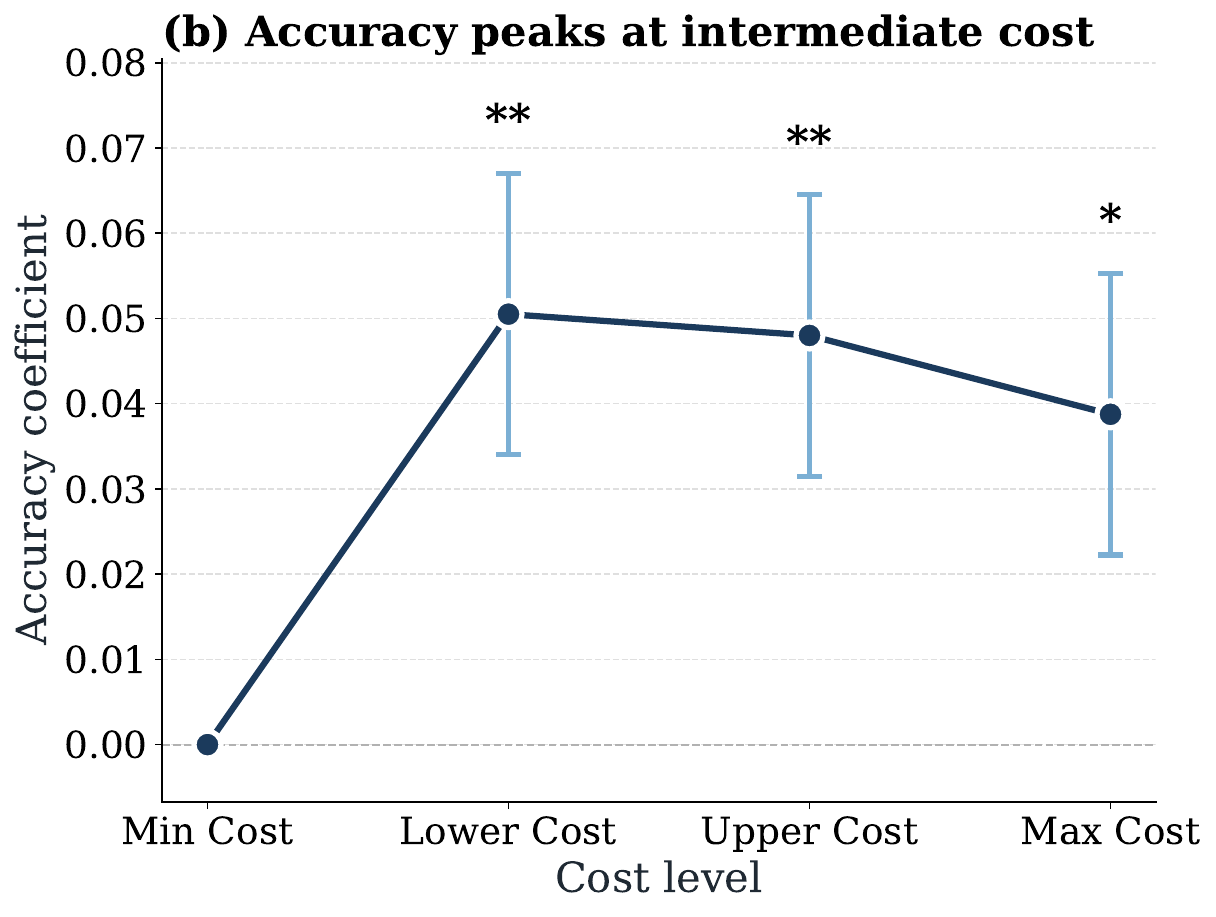}
    \label{fig:prompt-regression}
  \end{subfigure}
    \caption{
    \textbf{Task accuracy and token cost across models.}
    (a) Group-level accuracy and mean input tokens for each difficulty/model bin; the dashed line shows the overall trend.
    (b) Relative agent accuracy across cost quartiles, compared to the minimum-cost setting and estimated via mixed-effects regression. When working on the same problem, agent performance peaks at the intermediate-cost run and then saturates with higher costs.
    }
  \label{fig:accuracy-usage-cost}
\end{figure*}

\begin{figure}[t]
  \centering
  \begin{subfigure}{0.47\textwidth}
    \centering
    \includegraphics[width=\linewidth]{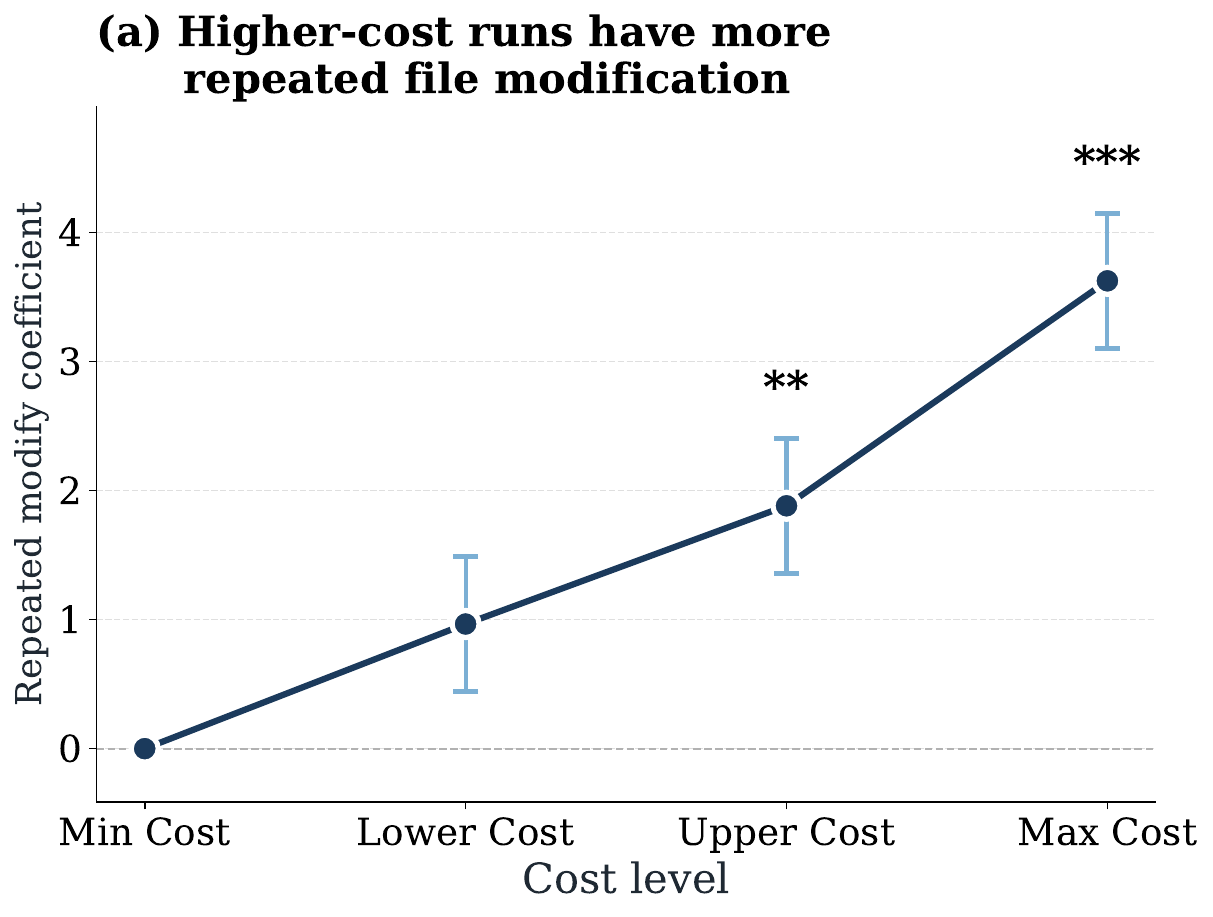}
    \label{fig:prompt-repeated-modify}
  \end{subfigure}
  \hfill
  \begin{subfigure}{0.47\textwidth}
    \centering
    \includegraphics[width=\linewidth]{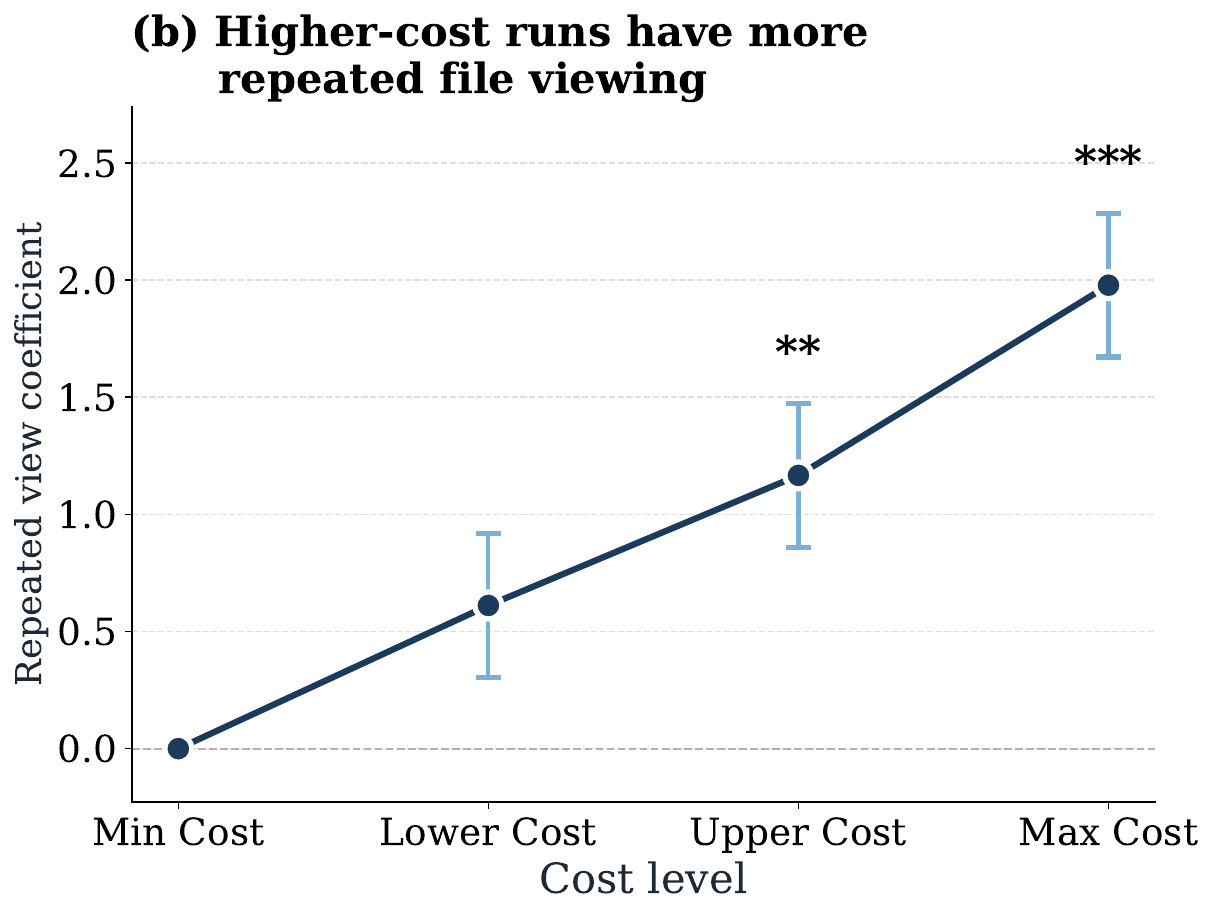}
    \label{fig:prompt-repeated-view}
  \end{subfigure}
    \caption{
    \textbf{When working on the same problem, high-cost runs are associated with repeated view and edits for the same file. }
    Relative frequency of repeated file modifications (a) and repeated file views (b) across cost quartiles, compared to the minimum-cost setting and estimated via mixed-effects regression.  
    }
  \label{fig:file-metrics-cost}
\end{figure}

Motivated by these observations, we further examine the behavioral patterns underlying high-cost failures by analyzing repeated file \emph{view} and \emph{modify} actions across cost levels.
As shown in Figure~\ref{fig:file-metrics-cost}, the frequency of repeated viewing and editing sharply increases in the more expensive runs. 
This indicates that many expensive but failed runs have potentially redundant back-and-forth file access and re-editing, reflecting inefficient search dynamics that inflate context length and token usage without proportional progress.
While not all high-cost runs are dominated by redundancy, this pattern provides a concrete behavioral explanation for the inverse accuracy--cost relationship observed above.

\begin{figure}[t]
    \centering
    \includegraphics[width=\textwidth]{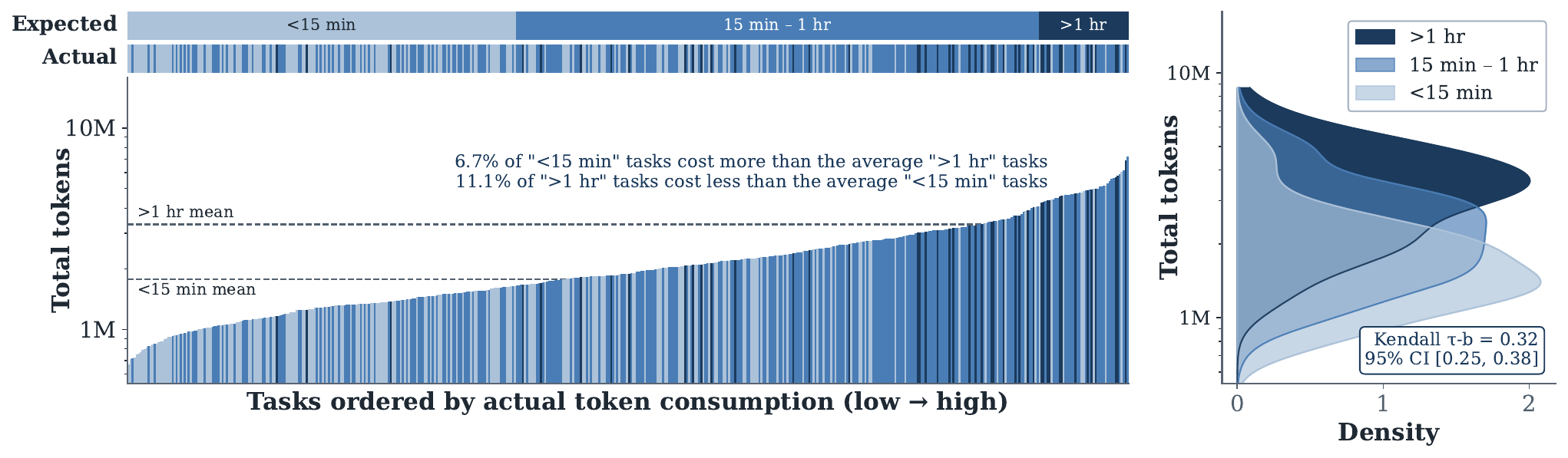}
    \caption{\textbf{Expert-rated task difficulty is a weak predictor of agent token consumption.} 
    Left: each vertical bar represents one of the 500 SWE-bench tasks, sorted by actual token consumption (low $\rightarrow$ high) and colored by human difficulty rating. The top reference strip shows the expected coloring under perfect alignment (clean light-to-dark gradient); the actual coloring below it is scrambled throughout. Dashed lines mark the $<$15 min and $>$1 hour group means.
    Right: density of token consumption per difficulty group, with substantial overlap across the full range. 
    }
    
    \label{fig:difficulty-metrics}
\end{figure}

\paragraph{Expert-rated task difficulty is a weak predictor of agent token consumption} 
Human engineers need different amounts of time and effort to complete different tasks. SWE-bench-Verified~\citep{jimenez2024swebench, chowdhury2024swebenchverified} provides expert-estimated difficulty levels which categorize problems based on the estimated time required by professional developers to resolve them (e.g., ``$<$15 min'', ``15 min -- 1 hour'', ``1--4 hours'', ``$>$4 hours''). Because there are only three instances in the $>$4 hours category, we merge it with the 1--4 hours group and report them together under ``$>$1 hour.''
Is expert-perceived task difficulty a good predictor of agent token usage?
Figure~\ref{fig:difficulty-metrics} shows the distribution of total token usage and compares it with the expected distribution if it were fully aligned with human-perceived difficulty. While overall resource consumption tends to rise with problem difficulty, the relationship is far from linear. The rank-monotonic association between human difficulty and token consumption is statistically real but modest (Kendall $\tau_b = 0.32$), and the two distributions overlap substantially: \textbf{6.7\%} of tasks labeled as ``$<$15-minute'' required more total tokens than the average ``$>$1-hour'' instance, and \textbf{11.1\%} of ``$>$1-hour'' tasks consumed fewer tokens than the average ``$<$15-minute'' instance.
These outliers highlight that human-estimated difficulty does not always align with the model's notion of complexity. Tasks that seem easy to humans may still demand extensive reasoning, exploration, or tool interaction from the model, whereas some ``hard'' problems may be efficiently solvable given the model's prior knowledge or search strategies. Consequently, human-labeled difficulty is a weak predictor of agent resource expenditure.

\section{Which Models are More Token Efficient}

The rapid growth in token consumption creates an urgent cost control problem for users and organizations deploying coding agents at scale. A natural response is to select the most cost-efficient model that can still complete the task. In this section, we examine the accuracy-cost trade-off and the token efficiency of eight frontier models.

\begin{figure}[t]
  \centering
  \begin{subfigure}[t]{0.47\textwidth}
  \label{fig:sec4-acc-token}
    \centering
    \includegraphics[width=\linewidth]{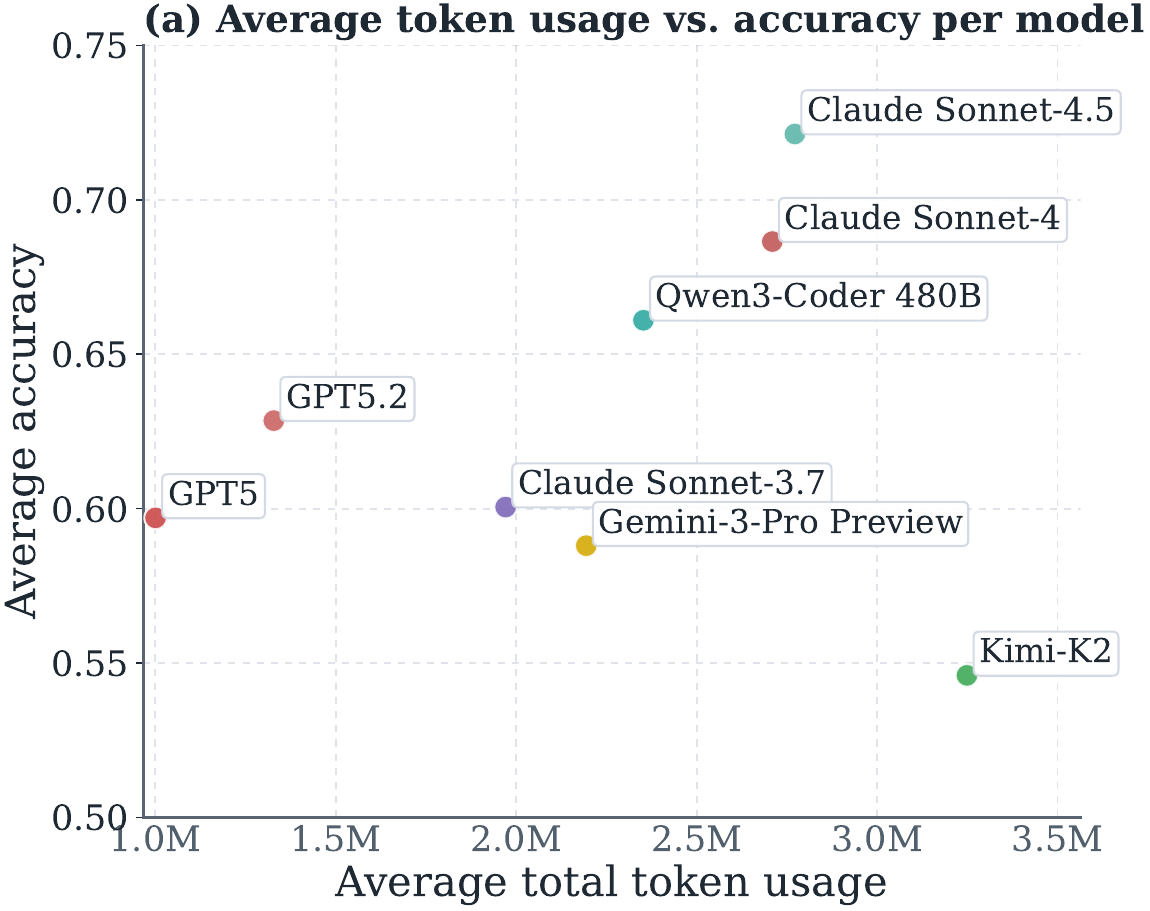}
  \end{subfigure}
  \hfill
  \begin{subfigure}[t]{0.47\textwidth}
  \label{fig:sec4-token-dev}
    \centering
    \includegraphics[width=\linewidth]{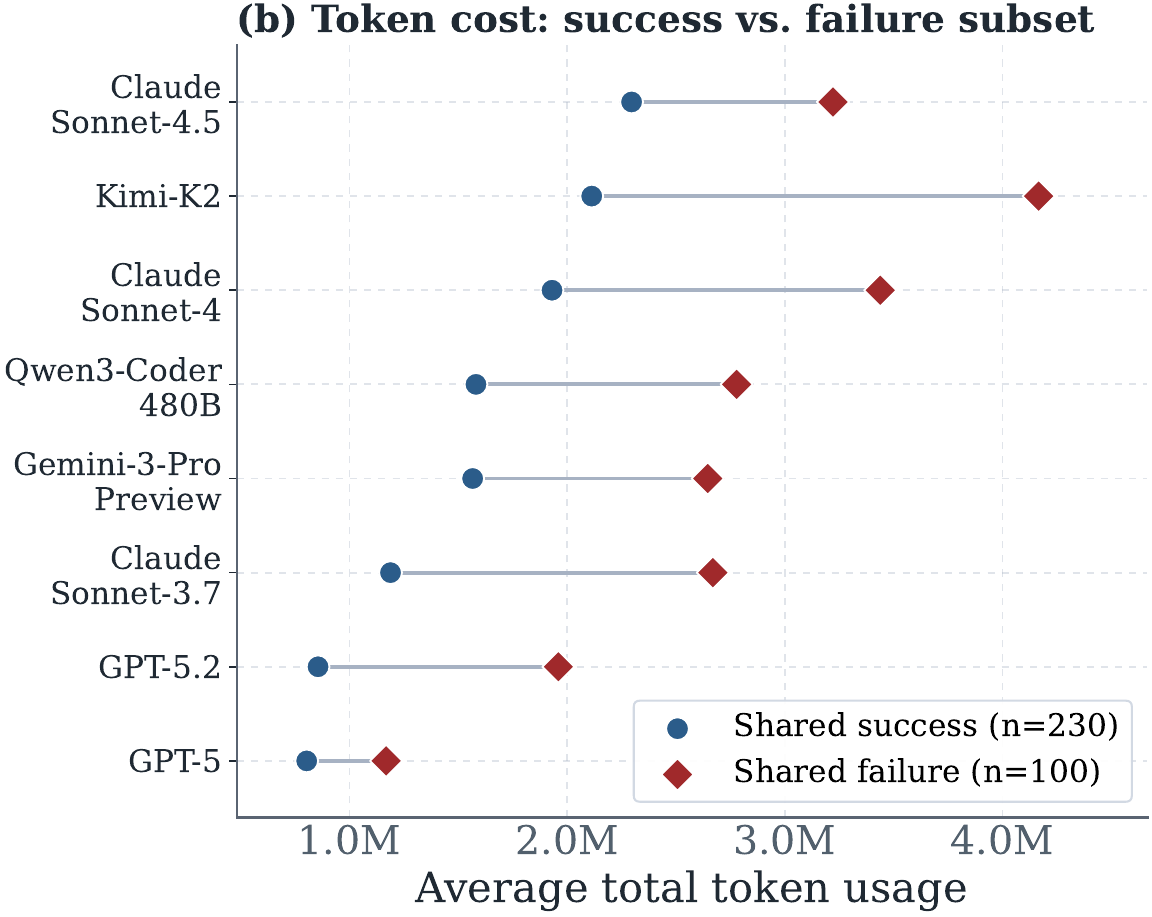}
  \end{subfigure}
  \caption{\textbf{Token efficiency varies substantially across models and reflects model-specific behavior rather than task difficulty.}
  (a) Mean total token usage vs.\ mean accuracy across all 500 SWE-bench instances; one point per model.
  (b) Model token usage on the shared success and failure tasks, blue dots show mean token usage on the shared success subset (n=230, solved by all models) and red diamonds show the shared failure subset (n=100, failed by all models). Models' relative token usage remains consistent on both subsets, suggesting that token efficiency is an inherent characteristic of the model.}
  \label{fig:sec4-efficiency}
\end{figure}

\paragraph{Accuracy--cost differences across models.}
To understand how token efficiency differs across models, we first analyze the trade-off between task accuracy and token consumption.
Figure~\ref{fig:sec4-efficiency}a shows that models operating at higher token budgets generally achieve higher accuracy, but vary significantly in how well they navigate this trade-off. GPT-5 and GPT-5.2 achieve strong accuracy at low cost, while Claude Sonnet~4.5, Claude Sonnet~4, and Qwen3-Coder-480B operate in a higher-cost regime. Kimi-K2 remains an outlier with both the highest cost and the lowest accuracy. Figure~\ref{fig:sec4-efficiency}b further shows the models' token usage across the shared success and shared failure subsets. In an ideal situation, stronger models should be able to consume fewer tokens on the easy task that all the models could solve and stop early on the hard tasks that every model fails. However, as shown in Figure~\ref{fig:sec4-efficiency}b, the relative ranking of models token usage persists across both the shared success and shared failure subsets. 
This suggests that the gap is not driven by task difficulty or by some models attempting harder problems. Instead, the same task is simply more expensive for some models than others, reflecting a behavioral tendency of the model rather than a property of the problem. Additionally, models generally consume more tokens on the shared failure subset than on the shared success subset, but the size of this gap varies substantially across models: GPT-5 and GPT-5.2 show only a mild increase ($<$0.5M tokens), whereas Kimi-K2 exhibits a much larger rise ($\sim$2M tokens). One likely explanation is that models lack a reliable mechanism to recognize when a task is unsolvable and stop early. Instead, they continue exploring, retrying, and re-reading context, accumulating cost without progress. The size of this excess spending appears to be model-specific, indicating that efficiency differences are systematic and become amplified under failure.

\begin{figure}[t]
  \centering
  \includegraphics[width=0.80\textwidth]{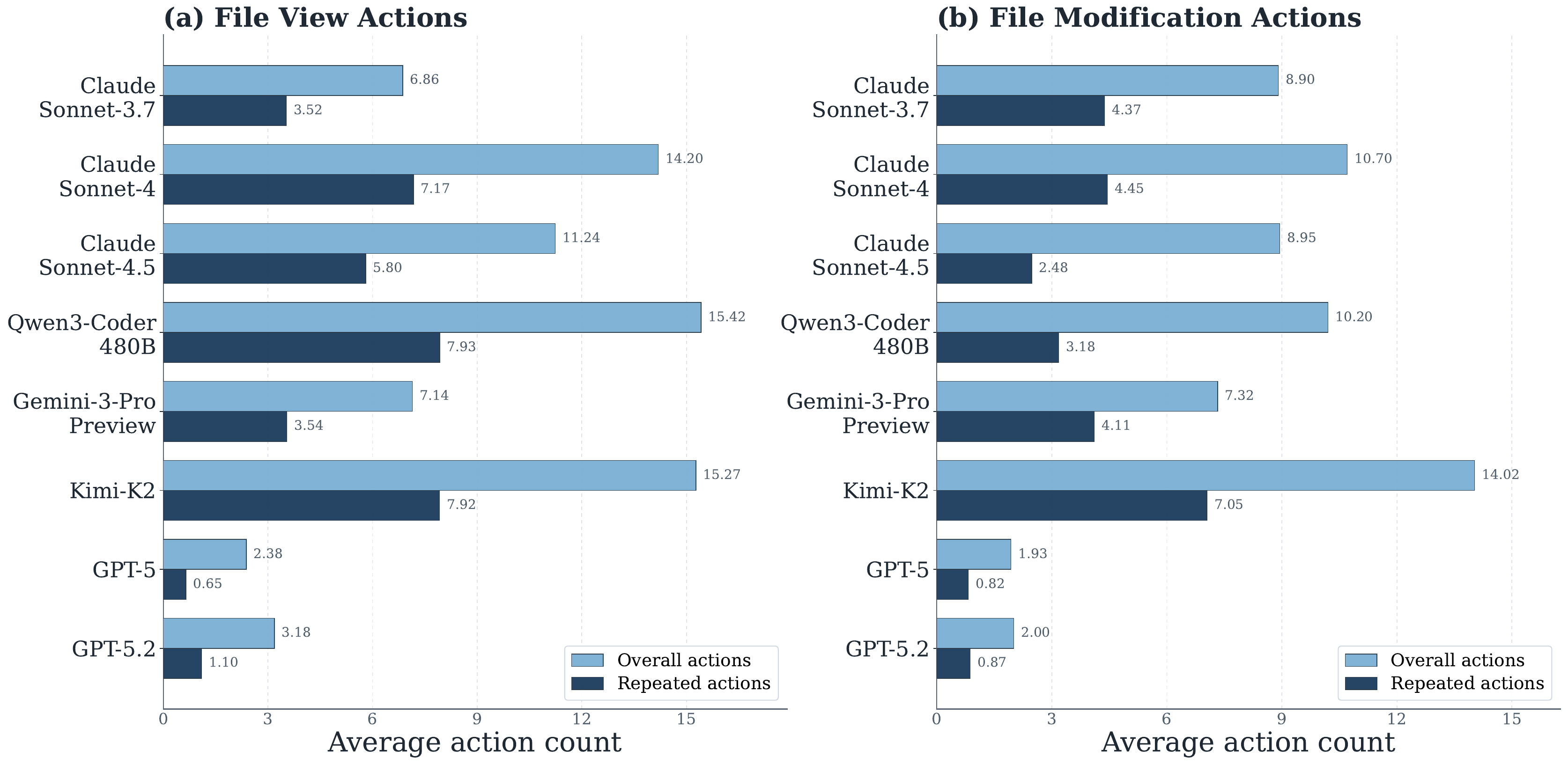}
  \caption{\textbf{Fine-grained file interaction patterns on the shared success subset.}
  For each model, we report the average number of overall and repeated file \emph{view} actions (a) and \emph{modification} actions (b). }
  \label{fig:sec4-actions}
\end{figure}

\paragraph{Fine-grained action differences across models.}
Building on the observed model-level differences in accuracy and token efficiency, we then examine more fine-grained action patterns to understand how these differences arise. More specifically, we look at file view actions and modification actions. As shown in Figure~\ref{fig:sec4-actions}, token-efficient models (GPT-5, GPT-5.2) perform fewer file views and modifications and have fewer repeated file actions. In contrast, higher-cost models such as Qwen3-Coder-480B, Claude Sonnet~4, and Kimi-K2 perform more file actions and around 50\% of them are repeated actions on the same file, indicating more exploratory and redundant interaction patterns.
Overall, these results indicate that model efficiency depends not only on the number of actions taken, but also on how effectively those actions are executed.

\section{Token–Cost Dynamics Across Phases and Rounds}
Long-horizon agentic tasks produce long and complex trajectories: the agent accumulates context over many rounds, interleaves different action types, and repeatedly reads and writes to the same files. On the cost side, LLM providers bill different token types at different rates. As a result, the same total token count can translate into very different monetary costs depending on when and how each token was spent.
In this section, we conduct a case study of Claude Sonnet-4.5 to open up the black box of agent cost and examine where tokens are spent and how that spending translates into dollars.

\subsection{Experimentation Setup}
Most commercial LLM providers charge different types of tokens at different rates. Output tokens are the most expensive because the model has to generate them one at a time. Standard input tokens, used when the model processes a fresh prompt, sit in the middle. Cached input tokens are the cheapest: once a chunk of context has been seen, the provider can reuse the work of processing it, so later reads are billed at a steep discount. Providers that expose caching explicitly, such as Anthropic, split the cache side further into \emph{cache creation} (writing context into the cache for later reuse) and \emph{cache read} (retrieving it in a later round at the discounted rate). This structure matters especially for agentic workloads: because long trajectories keep adding to the context, the same material would otherwise be re-processed on every round.  Caching, therefore, has become an essential strategy for keeping token costs manageable. 

We analyze agent cost at two levels of granularity. The first is the \emph{phase} level, where we divide each trajectory into five stages of problem-solving (Setup, Explore, Fix, Validate, Closeout) and compare how token counts and costs vary from stage to stage. The second is the \emph{round} level, where we trace a single trajectory step by step to see how each round's cost breaks down by token type and which type dominates at different points in the task. We use Claude Sonnet-4.5 for the analyses in this section. Its API reports each round's cost as the sum of four separately-priced categories: non-cached input, output, cache creation, and cache read. Cache creation is priced according to how long the cache persists; we use the 5-minute write rate throughout.\footnote{Pricing details at \url{https://docs.claude.com/en/docs/about-claude/pricing}; full cost formulas in Appendix~\ref{app:cost-formulas}.}

\subsection{Phase-Level Token Usage Dynamics}
We divide each problem-solving trajectory into five semantically grounded phases based on the agent's functional behavior: \emph{Setup}, \emph{Explore}, \emph{Fix}, \emph{Validate}, and \emph{Closeout}. Table~\ref{tab:phase-breakdown} describes each phase and reports its share of total rounds. The Fix and Explore phases together account for roughly two-thirds of all rounds, while Setup, Validate, and Closeout fill out the remaining third. For each phase, we aggregate across 500 problem instances and compute average token counts, dollar costs, and per-phase correlations between token types and total cost.

\begin{table}[t]
\centering
\small
\begin{tabular}{l l c}
\toprule
\textbf{Phase} & \textbf{Description} & \textbf{Proportion} \\
\midrule
Setup & Task planning, environment setup, initial reproduction & 9.98\% \\
Explore & Code search, file inspection, root-cause analysis & 30.37\% \\
Fix & Code edits, debugging iterations, patch refinement & 33.53\% \\
Validate & Testing, regression checks, verification & 16.59\% \\
Closeout & Final checks, cleanup, summary output & 9.53\% \\
\bottomrule
\end{tabular}
\caption{Phases of agent trajectories. Percentages are computed over all rounds across Sonnet-4.5 runs.}
\label{tab:phase-breakdown}
\end{table}
\vspace{-0.5cm}

\noindent

\begin{figure}[t]
  \centering
  \begin{subfigure}{0.4\linewidth}
    \centering
    \includegraphics[width=\linewidth]{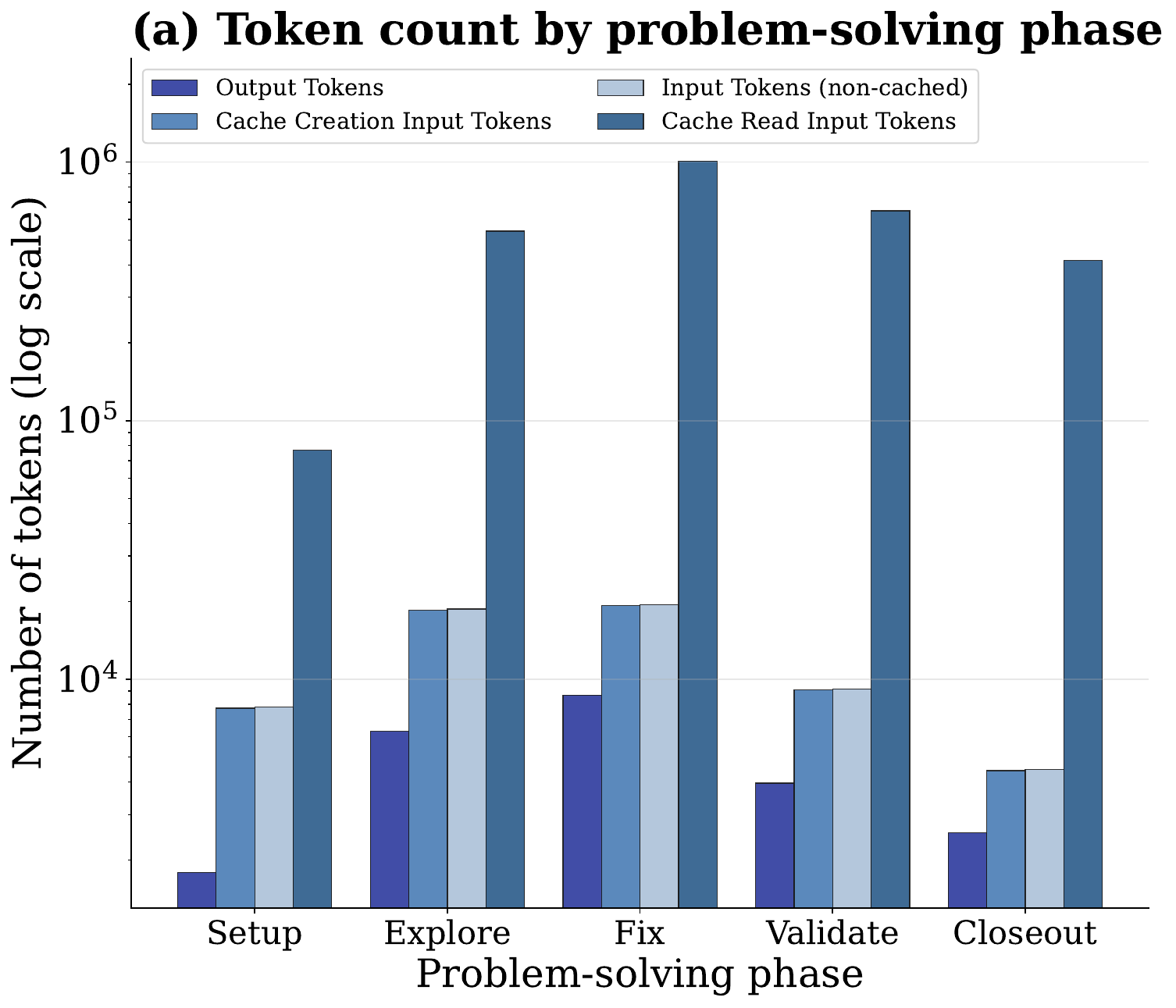}
  \end{subfigure}
  \begin{subfigure}{0.4\linewidth}
    \centering
    \includegraphics[width=\linewidth]{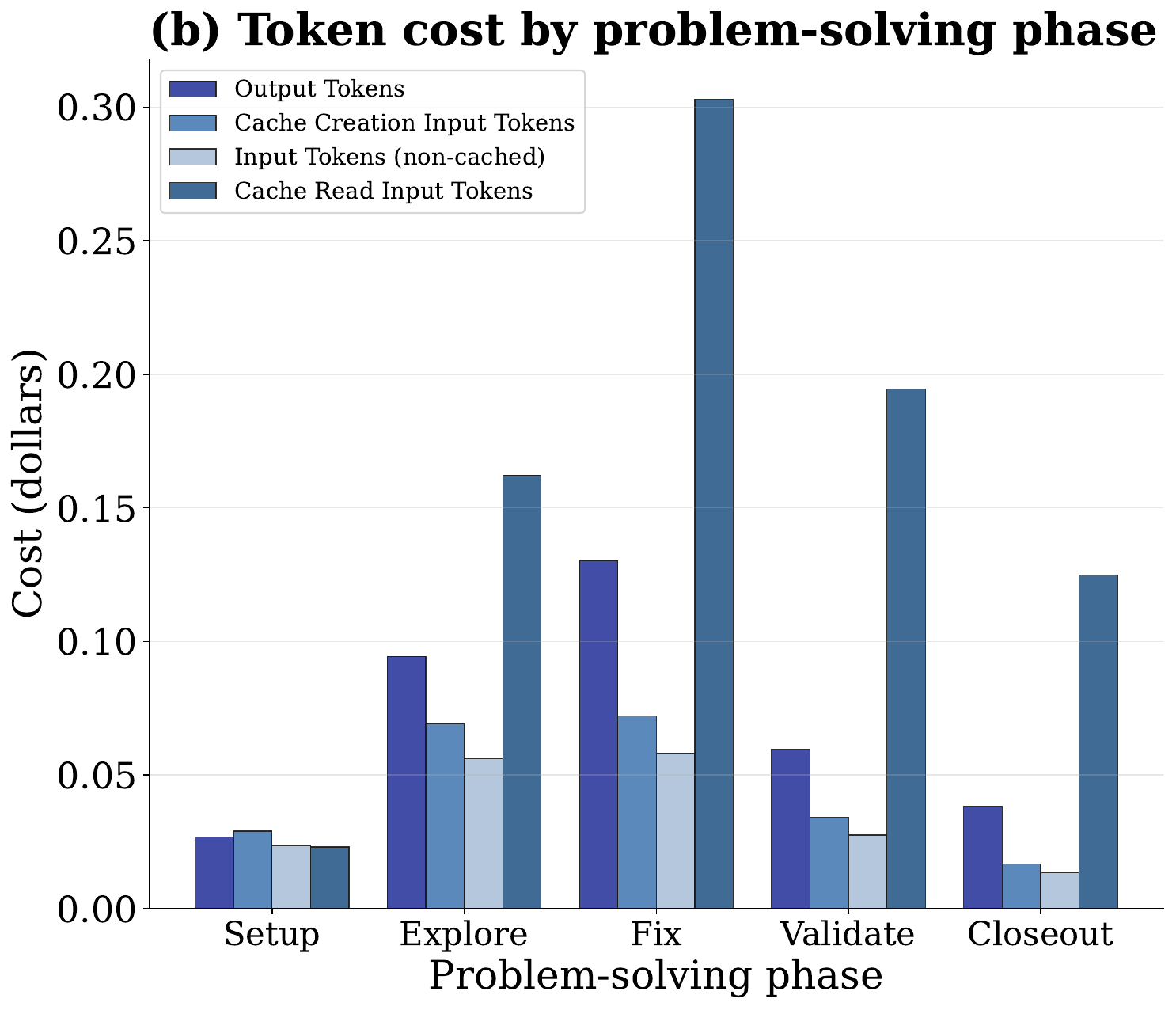}
  \end{subfigure}
    \caption{\textbf{Phase-level token usage and cost dynamics.} Input tokens dominate both raw token usage and dollar cost across phases.}
    
  \label{fig:cache-correlation-three}
\end{figure}

We find that \textbf{cache reads dominate both raw token volume and dollar cost.}
In every phase, cache-read input tokens are the largest category by a wide margin (Figure~\ref{fig:cache-correlation-three}a), reflecting the cumulative reuse of prior context. Non-cached input and cache-creation tokens track each other closely, consistent with newly introduced context being cached as soon as it enters the conversation, while output tokens are relatively high only in \emph{Setup}, where planning-heavy generation is concentrated. Figure~\ref{fig:cache-correlation-three}b further shows the actual dollar cost of each stage.  Cache reads remain the dominant cost contributor in every phase. Given that output tokens are individually priced roughly $80\times$ higher than cache reads, such a result further highlights that the sheer volume of accumulated context is large enough that cheap-per-token cache reads still outweigh expensive-per-token output in aggregate.


\subsection{Round-Level Cost Dynamics}
\begin{figure*}[t]
  \centering
  \includegraphics[width=0.95\textwidth]{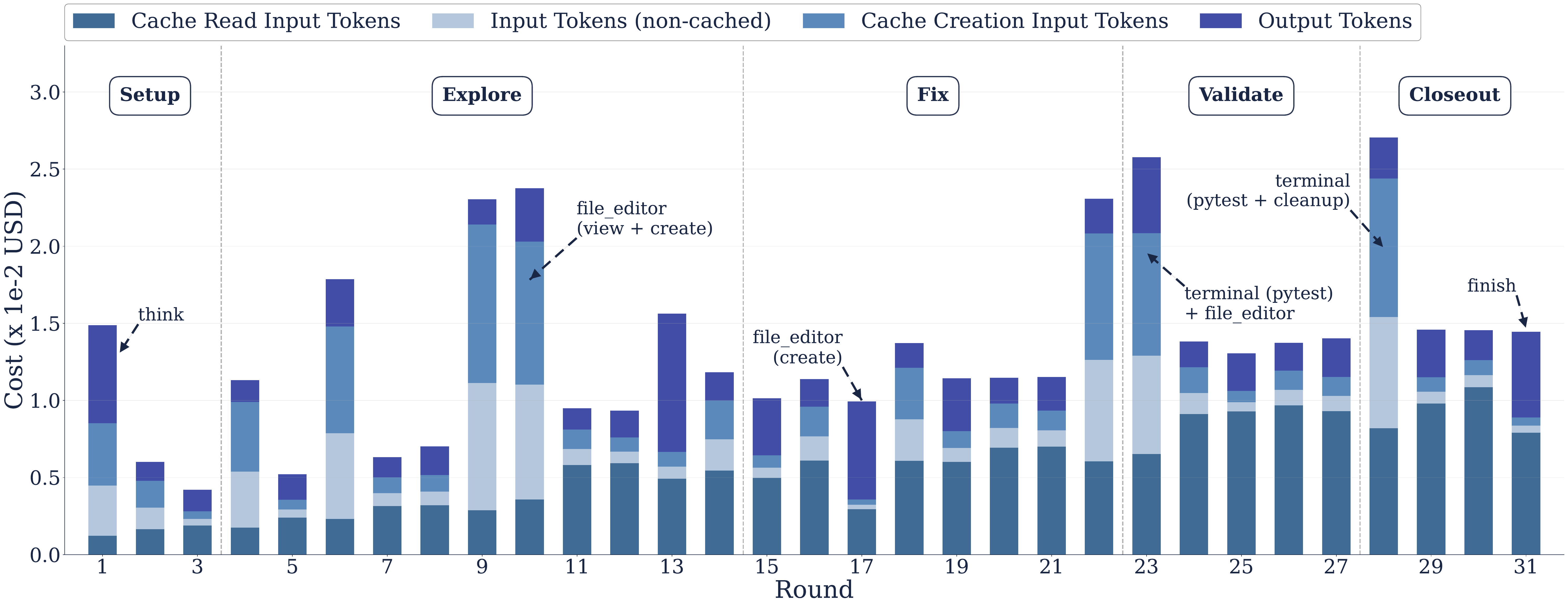}
  \caption{\textbf{Round-level token-cost dynamics of the agent trajectory on \textit{astropy\_\_astropy-7336}.}
  The cost of cache read stably increases with the accumulation of the input contexts. Cost spikes are driven by discrete actions that introduce new content (file views, test execution, script generation, final summary).}
  \label{fig:round-cost-breakdown}
\end{figure*}

To further illustrate the round-level token-cost dynamics, we zoom into a representative agent trajectory (on task \textit{astropy\_\_astropy-7336}). Figure~\ref{fig:round-cost-breakdown} shows the per-round cost decomposed across the four token categories. 

Cache-read costs accumulate gradually as the trajectory progresses and form a relatively stable baseline in each round. Total per-round cost, however, is far from monotonic. The visible cost spikes are driven by specific actions that introduce new content into the context: repository exploration, file creation, test execution, and final summarization. In other words, the accumulated cost of reusing context is steady and predictable; what makes individual rounds expensive is what the agent chooses to add to the context on that round.

These patterns line up with the functional roles of the different phases. In the \emph{Setup} phase, where the agent starts to reason and plan,  output tokens dominate the overall costs. Following, as repository inspection and code reading pull large amounts of content into the context window, the input tokens gradually take over in the \emph{Explore} phase. In the later phases (\emph{Fix}, \emph{Validate}, \emph{Closeout}), output tokens come back into play for script generation and code edits, while input tokens reflect the cost of reading test results and execution output. Table~\ref{tab:traj} further illustrates representative rounds and their dominant cost source. Rounds 10, 23, and 28 involve input-heavy tool calls like viewing new files, running tests, and cleaning up artifacts, which lead to a sharp increase in non-cached input cost. As a comparison, rounds 1, 17, and 31 involve output-heavy actions like planning and editing, leading to a high output token cost. 


\begin{table*}[t]
\centering
\resizebox{\textwidth}{!}{
\renewcommand{\arraystretch}{0.9}
\begin{tabular}{
>{\centering\arraybackslash}m{1.2cm}
m{3.2cm}
m{5.5cm}
m{7.5cm}
}
\toprule
\textbf{Round} & \textbf{Dominant cost source} & \textbf{Tool usage} & \textbf{Action summary} \\
\midrule
1  & Output & \texttt{think} & Planning and reasoning about the issue. \\
10 & Non-cached input & \texttt{file\_editor} (view + create) & Reads test file and creates reproducer. \\
17 & Output & \texttt{file\_editor} (create) & Writes test script for debugging. \\
23 & Non-cached input & \texttt{terminal} (pytest) + \texttt{file\_editor} & Runs tests and creates verification script. \\
28 & Non-cached input & \texttt{terminal} (pytest + cleanup) & Runs tests and cleans up files. \\
31 & Output & \texttt{finish} & Produces final summary of the fix. \\
\bottomrule
\end{tabular}
}
\caption{Six representative rounds from the trajectory in Figure~\ref{fig:round-cost-breakdown}, grouped by their dominant cost source. Input-heavy rounds are driven by new content returned from tool calls; output-heavy rounds are driven by the agent's own reasoning and generation.}
\label{tab:traj}
\end{table*}

\section{Predicting Agent Token Consumption before Execution}
The patterns discussed so far expose a fundamental tension in how AI agents are priced today. Token consumption varies widely across tasks and runs, and higher cost does not reliably translate into better outcomes. Users end up committing to a bill they cannot see in advance, sometimes paying substantial sums for runs that ultimately fail. Providers face a different but related problem: without a way to anticipate cost up front, it is hard to design pricing tiers that feel predictable to customers while staying financially viable, and hard to enforce budget caps or catch expensive runs before they spiral. Reliable cost estimation \emph{before} execution would help on both sides. Users could compare agents on expected cost rather than hoping for the best. Providers could offer tiered plans and budget guarantees with known exposure. And both could set early alerts on runs heading toward the high tail of the cost distribution.

In this section, we set up the \textbf{agent token consumption prediction} problem and empirically test whether agents can predict their own token costs before executing a task. We focus on \emph{self-prediction}, where the same coding agent used for task execution is repurposed to estimate token usage. We view this as a foundational capability for autonomous agents: an agent that can reason about its own behavior well enough to anticipate the resources it will consume is also better positioned to plan, budget, decide when a task is worth attempting, and recognize when to stop. Cost estimation is one concrete instance of this broader capacity for behavioral self-modeling, and one that is directly measurable. Beyond this, self-prediction is an appealing setting for two practical reasons. First, the executing agent has privileged access to the information that drives cost: the repository structure it would explore, the tools it would call, and the planning depth it would invoke. A separate predictor would have to reconstruct this context from scratch. Second, self-prediction requires no additional model, training pipeline, or infrastructure to deploy: any agent that can run a task can, in principle, also estimate its cost, making the approach immediately usable in existing systems.

\paragraph{Experimentation settings} We use the coding agent itself as the predictor of its own token consumption. The agent retains its full tool-calling and interaction capabilities, allowing it to inspect the repository structure, run preliminary commands, and reason about potential execution paths before producing an estimate, but is instructed to output a token estimate rather than attempt a fix. This mirrors how a developer might inspect a codebase to scope effort before committing to an implementation. Prior work on self-feedback~\citep{madaan2023self} and language model calibration~\citep{kadavath2022language} provides theoretical grounding for the broader claim that models can reason about their own outputs and uncertainty.
We use a fine-grained prediction prompt: the agent is instructed to decompose the task into stages and report separate estimates for input tokens, output tokens, and total cost (see Appendix~\ref{prompt:agent-zero-shot-grad-prompt}). The prompt also includes one human-written worked example that demonstrates the expected reasoning process and output format. Due to budget constraints, we run three independent predictions per model on the same 500 SWE-bench instances. We evaluate prediction quality with Pearson correlation between predicted and actual token counts, and we additionally report the \emph{overhead} of self-prediction as the ratio of prediction cost to actual task cost.

\paragraph{Results}

\begin{figure}[t]
  \centering
  \hspace*{-1.2cm}
  \includegraphics[width=0.80\linewidth]{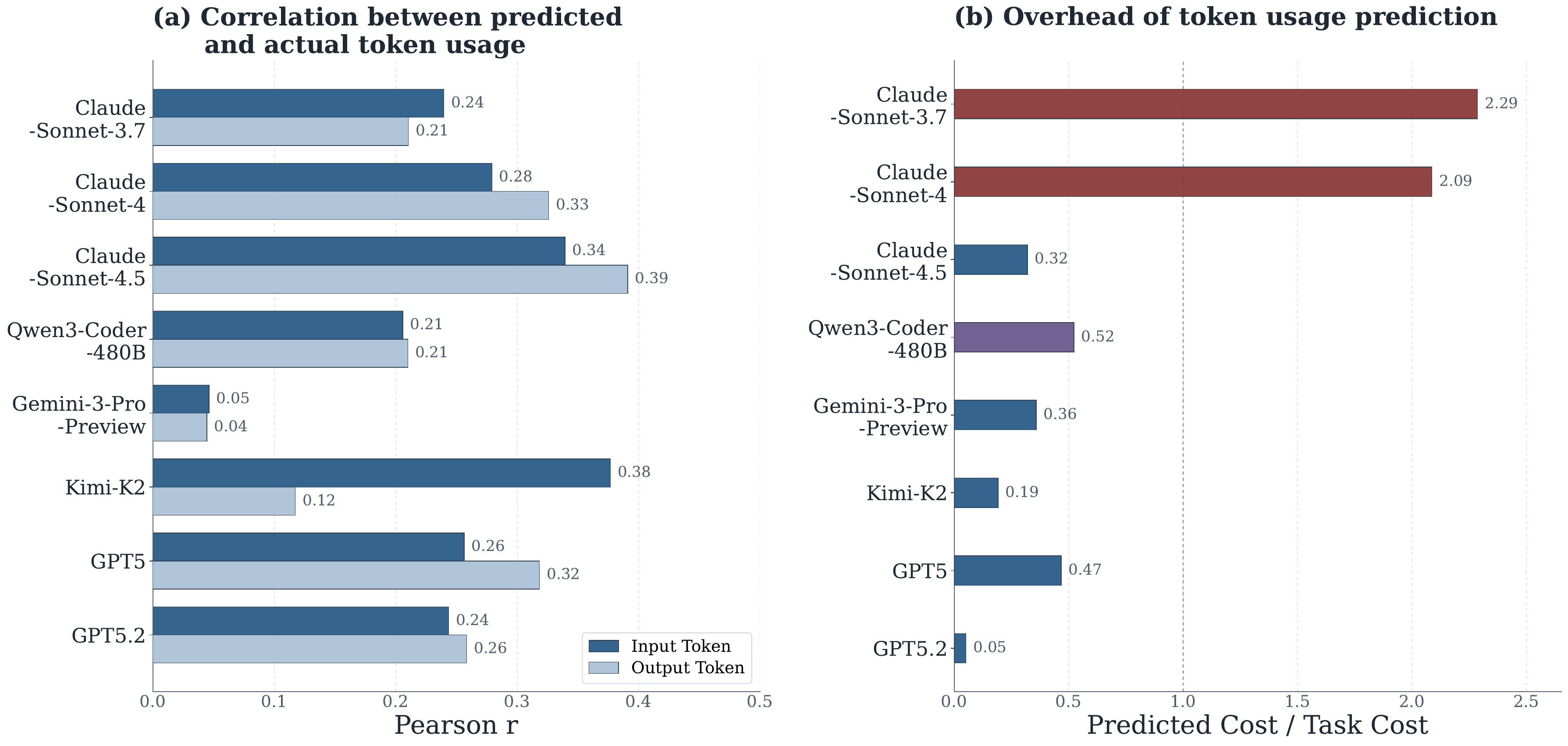}
  \caption{\textbf{Self-prediction performance and overhead across models.}
  (a) Pearson correlation between predicted and actual token counts. 
  (b) Overhead of self-prediction, measured as the ratio of prediction cost to actual task cost. Overall, predicting token usage before execution is challenging for all the tested models and there is much space for improving the prediction efficiency.}
  \label{fig:prediction-correlation}
\end{figure}

Figure~\ref{fig:prediction-correlation} summarizes prediction performance and overhead across the eight models. Overall, self-prediction achieves non-trivial but modest correlations with real token usage overall. Within the Claude Sonnet family, correlation improves steadily with newer generations and peaks at 0.39 for output-token prediction with Sonnet~4.5. GPT-5, GPT-5.2, Kimi~K2, and Qwen3-Coder reach similar modest correlations, while Gemini-3-Pro trails well behind on both input and output tokens.  Input-token prediction is consistently harder than output-token prediction, which is unsurprising given the scale and growth rate of input tokens over long trajectories. Kimi~K2 is the one exception: it attains the highest input-token correlation (0.38), suggesting it is potentially more sensitive to context expansion than the other models. Taken together, these results indicate that self-prediction captures coarse-grained trends in token usage but remains noisy at the instance level.

\begin{figure*}[t]
  \centering
  \begin{subfigure}{0.49\textwidth}
    \centering
    \includegraphics[width=\linewidth]{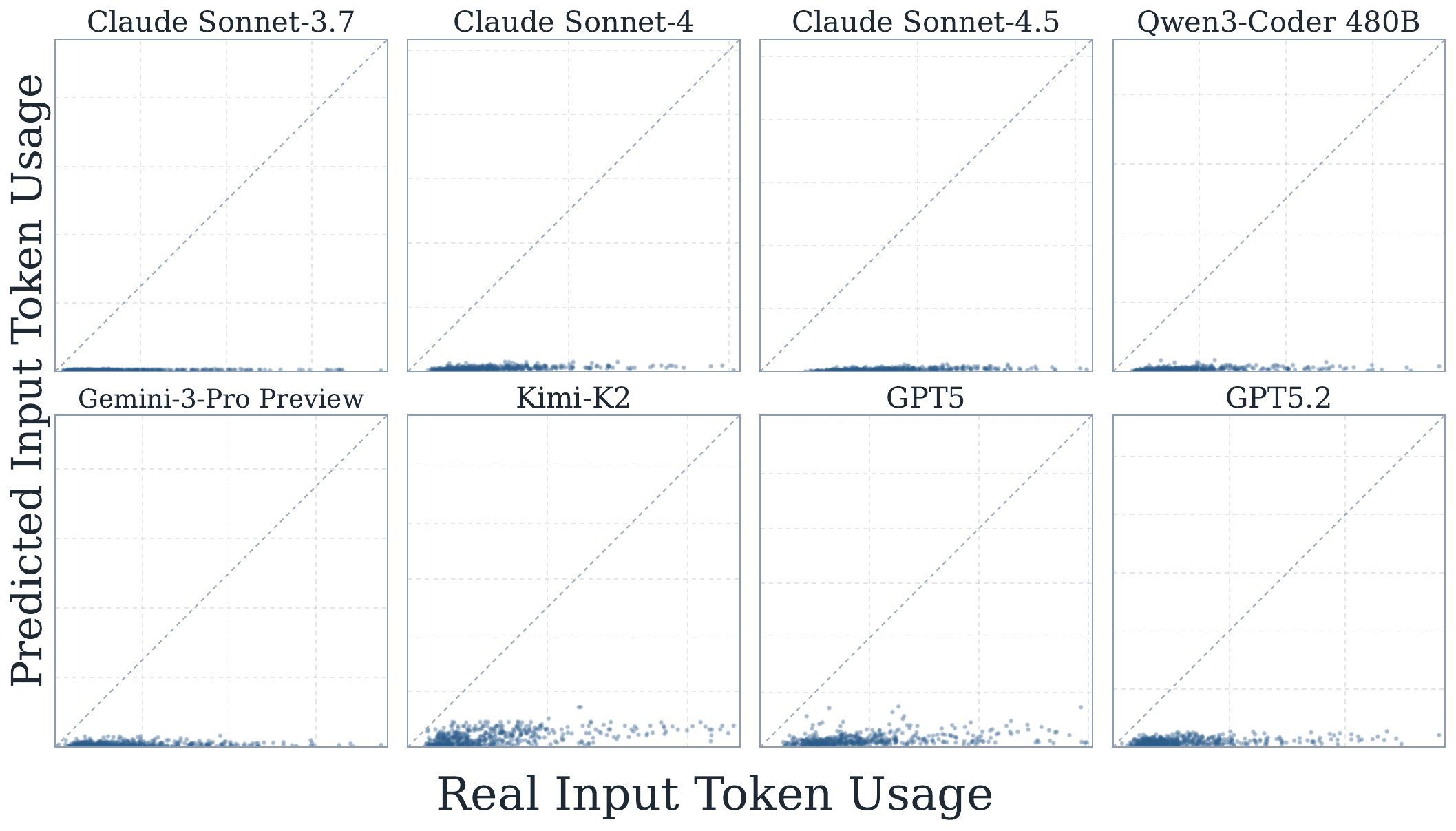}
  \end{subfigure}
  \hfill
  \begin{subfigure}{0.49\textwidth}
    \centering
    \includegraphics[width=\linewidth]{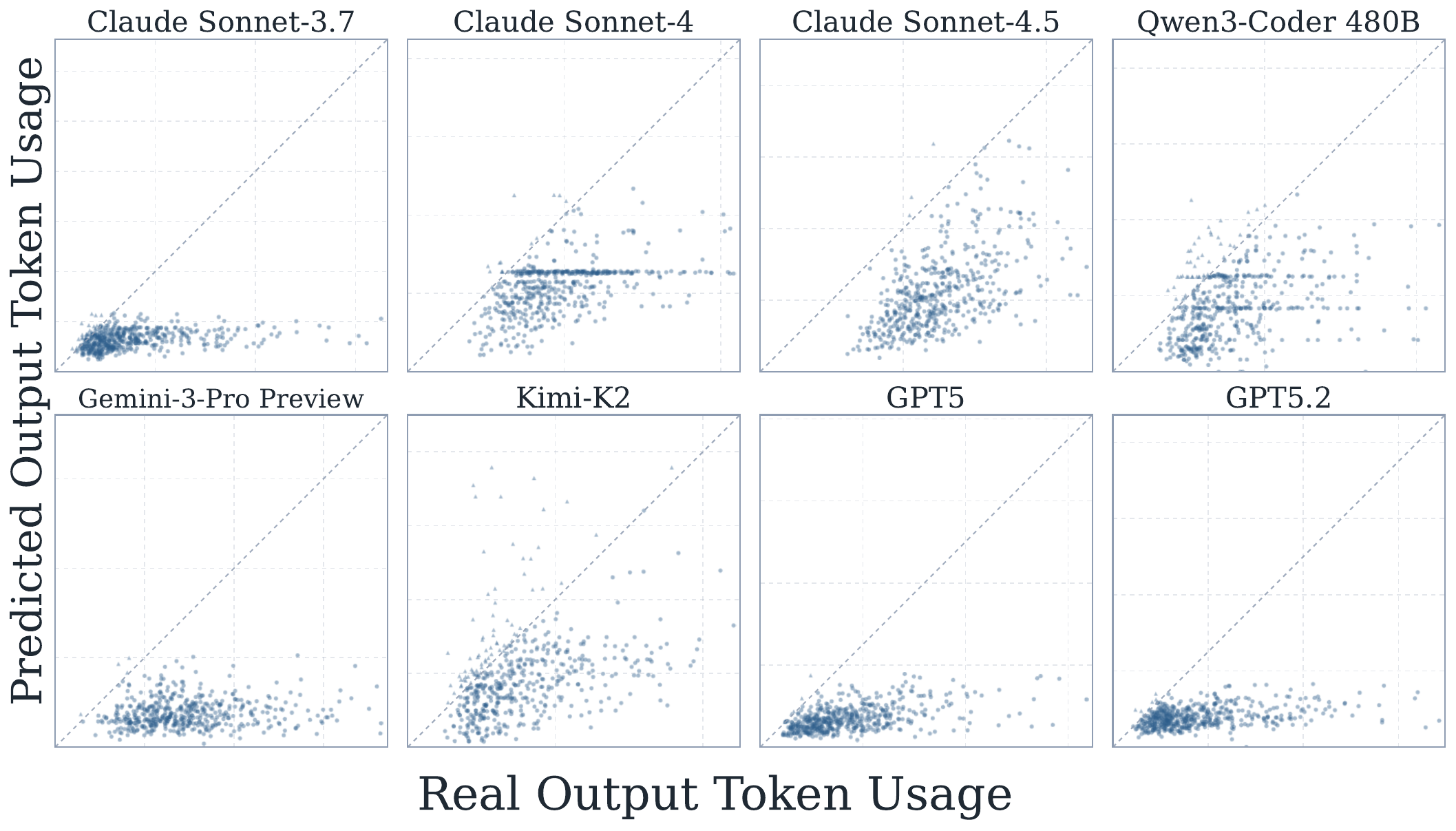}
  \end{subfigure}
  \caption{\textbf{Predicted vs.\ real token usage across models.} The dashed diagonal indicates perfect calibration. Agents systematically underestimate both input and output token usage.}
  \label{fig:prediction-scatter}
\end{figure*}

\paragraph{Prediction overhead relative to task execution} Given that token consumption prediction is also an agentic task, the prediction itself may lead to additional token costs. In an ideal situation, we define the prediction overhead as the ratio of prediction token cost to actual task token cost and present the result in Figure~\ref{fig:prediction-correlation}b. For most models, self-prediction is substantially cheaper than execution itself, typically costing less than half the original task. But the relationship between overhead and accuracy is not monotonic. Sonnet~3.7 and Sonnet~4 spend more than $2\times$ the task cost on prediction and yet do not achieve the strongest correlations. Sonnet~4.5 delivers the highest correlation at just $0.32\times$ the task cost, and GPT-5.2 drops prediction overhead below 6\% while still hitting moderate correlations. Taken together, these numbers suggest that better prediction is possible with a reasonable amount of compute, and there is substantial room to improve prediction accuracy without proportional increases in overhead.

\paragraph{Models systematically underestimate the tokens they need.}
Correlation captures the strength of the association between predicted and actual token usage but not its direction. To examine whether models over- or underestimate, we compare the predicted and actual token distributions in Figure~\ref{fig:prediction-scatter}. Models consistently underestimate the tokens they need: most points fall below the diagonal for every model we tested. The bias is especially pronounced for input tokens, whose predictions stay compressed even as real values grow into the millions. 
Appendix~\ref{sec:no-icl} provide further evidence that this pattern persist when no in-context example is presented.

Taken together, our results indicate that predicting token usage before execution is a genuinely difficult task for current models. Correlations with actual usage are consistently above chance but remain too modest to support precise, instance-level cost estimates. Self-prediction also carries non-trivial latency and overhead of its own, especially for models that explore extensively before committing to a number, which is hard to justify in real interactive or time-sensitive settings. Our result suggest that while self-prediction could still be useful as a coarse-grained signal of relative cost and task difficulty, making it reliable, efficient, and cleanly integrated with execution remains an open problem.

\section{Discussion}
In this paper, we presented the first systematic analysis of agent token consumption and empirically tested whether models can predict their own token cost before execution. In this section, we discuss the main limitations of our study and the implications of our findings for the design and pricing of agent-based systems.

\paragraph{Limitations}
One of the key limitations of our study is the set of agentic models we evaluate. We cover eight frontier models (Claude Sonnet~3.7, Sonnet~4, Sonnet~4.5, GPT-5, GPT-5.2, Qwen3-Coder, Kimi-K2, and Gemini~3 Pro Preview), which is a broad sample by the standards of existing work, but still only a slice of the agentic model landscape. Collecting full execution trajectories is computational expensive, which constrained how many models we could include. The qualitative patterns we observe hold consistently across the models we tested, but validating them on a wider range of architectures and agent designs would further strengthen their generality. To support such extensions, we release our experimental pipeline so that future work can replicate and build on our analysis.

\paragraph{User transparency}
Reliable token usage predictions before execution is very important for greater pricing transparency and user trust. Ideally, an agentic system would tell users the likely cost of a task before execution, letting them make informed decisions. Current language models are not yet good enough at point estimation to make this realistic for exact costs. Our results nonetheless suggest that agents themselves can potentially serve as useful predictors of their own cost, at least at the coarse-grained level of identifying high-cost tasks. Even without precise estimates, this kind of signal is enough for providers to issue early warnings, request explicit user approval, or offer alternative execution modes before committing to an expensive run.

\paragraph{Agent pricing}
Pricing is one of the central challenges for providers of agentic systems. Subscription models work for products like ChatGPT because typical users consume a predictable, bounded number of tokens. Agentic tasks break this assumption: even simple problems can burn through large token budgets due to multi-step reasoning and tool use, which makes accurate cost prediction important for designing sustainable pricing strategies. Our findings show that token usage, especially input tokens, is highly variable and hard to predict because agent trajectories are inherently stochastic. Purely upfront pricing therefore remains difficult, and consumption-based pricing will likely stay the most practical option until pre-execution estimation becomes more reliable. Complementary mechanisms such as budget-aware tool-use policies~\citep{liu2025budget} can help mitigate cost volatility by ... enforcing runtime token constraints. More broadly, designing pricing schemes that are both sustainable for providers and predictable for users remains an important open direction for future research.

\section{Conclusion}
With the rapid growth of token consumption in agentic settings, the ability to predict token usage before a task executes becomes central to building transparent and sustainable pricing models for AI agents. In this paper, we present the first systematic study of agent token consumption and empirically evaluate whether models can predict their own token usage before execution. Our results suggest that agentic tasks lead to complex token usage dynamics and that predicting potential token consumption before task execution remains a fundamentally challenging problem for frontier models. Our study provides new insights on agent behavior and could inspire new studies on building more controllable and transparent agent pricing schemes.

\newpage

\bibliography{colm2026_conference}

\begin{thebibliography}{25}
\providecommand{\natexlab}[1]{#1}
\providecommand{\url}[1]{\texttt{#1}}
\expandafter\ifx\csname urlstyle\endcsname\relax
  \providecommand{\doi}[1]{doi: #1}\else
  \providecommand{\doi}{doi: \begingroup \urlstyle{rm}\Url}\fi

\bibitem[Aggarwal et~al.(2025)Aggarwal, Kim, Lanchantin, Welleck, Weston, Kulikov, and Saha]{aggarwal2025optimalthinkingbench}
Pranjal Aggarwal, Seungone Kim, Jack Lanchantin, Sean Welleck, Jason Weston, Ilia Kulikov, and Swarnadeep Saha.
\newblock Optimalthinkingbench: Evaluating over and underthinking in llms.
\newblock \emph{arXiv preprint arXiv:2508.13141}, 2025.

\bibitem[Chen et~al.(2026)Chen, Zhang, He, Stoica, Zaharia, and Zou]{chen2026price}
Lingjiao Chen, Chi Zhang, Yeye He, Ion Stoica, Matei Zaharia, and James Zou.
\newblock The price reversal phenomenon: When cheaper reasoning models end up costing more.
\newblock \emph{arXiv preprint arXiv:2603.23971}, 2026.

\bibitem[Chowdhury et~al.(2024)Chowdhury, Aung, Shern, Jaffe, Sherburn, Starace, Mays, Dias, Aljubeh, Glaese, Jimenez, Yang, Ho, Patwardhan, Liu, and Madry]{chowdhury2024swebenchverified}
Neil Chowdhury, James Aung, Chan~Jun Shern, Oliver Jaffe, Dane Sherburn, Giulio Starace, Evan Mays, Rachel Dias, Marwan Aljubeh, Mia Glaese, Carlos~E. Jimenez, John Yang, Leyton Ho, Tejal Patwardhan, Kevin Liu, and Aleksander Madry.
\newblock Introducing {SWE}-bench verified, 2024.
\newblock URL \url{https://openai.com/index/introducing-swe-bench-verified/}.

\bibitem[{Crystalcare AI}(2023)]{codefeedback_sharegpt}
{Crystalcare AI}.
\newblock Code-feedback sharegpt.
\newblock \url{https://huggingface.co/datasets/Crystalcareai/Code-feedback-sharegpt-renamed}, 2023.
\newblock A large-scale dataset of coding-related multi-turn chat conversations derived from ShareGPT, focusing on code feedback and interactive code discussion.

\bibitem[Fan et~al.(2025)Fan, Vasilevski, Lin, Chen, Chen, Zhong, Zhang, He, and Hassan]{fan2025swe}
Zhiyu Fan, Kirill Vasilevski, Dayi Lin, Boyuan Chen, Yihao Chen, Zhiqing Zhong, Jie~M Zhang, Pinjia He, and Ahmed~E Hassan.
\newblock Swe-effi: Re-evaluating software ai agent system effectiveness under resource constraints.
\newblock \emph{arXiv preprint arXiv:2509.09853}, 2025.

\bibitem[Gema et~al.(2025)Gema, H{\"a}gele, Chen, Arditi, Goldman-Wetzler, Fraser-Taliente, Sleight, Petrini, Michael, Alex, et~al.]{gema2025inverse}
Aryo~Pradipta Gema, Alexander H{\"a}gele, Runjin Chen, Andy Arditi, Jacob Goldman-Wetzler, Kit Fraser-Taliente, Henry Sleight, Linda Petrini, Julian Michael, Beatrice Alex, et~al.
\newblock Inverse scaling in test-time compute.
\newblock \emph{arXiv preprint arXiv:2507.14417}, 2025.

\bibitem[Gu et~al.(2024)Gu, Roziere, Leather, Solar-Lezama, Synnaeve, and Wang]{pmlr-v235-gu24c}
Alex Gu, Baptiste Roziere, Hugh~James Leather, Armando Solar-Lezama, Gabriel Synnaeve, and Sida Wang.
\newblock {CRUXE}val: A benchmark for code reasoning, understanding and execution.
\newblock In Ruslan Salakhutdinov, Zico Kolter, Katherine Heller, Adrian Weller, Nuria Oliver, Jonathan Scarlett, and Felix Berkenkamp (eds.), \emph{Proceedings of the 41st International Conference on Machine Learning}, volume 235 of \emph{Proceedings of Machine Learning Research}, pp.\  16568--16621. PMLR, 21--27 Jul 2024.
\newblock URL \url{https://proceedings.mlr.press/v235/gu24c.html}.

\bibitem[Jimenez et~al.(2024)Jimenez, Yang, Wettig, Yao, Pei, Press, and Narasimhan]{jimenez2024swebench}
Carlos~E Jimenez, John Yang, Alexander Wettig, Shunyu Yao, Kexin Pei, Ofir Press, and Karthik~R Narasimhan.
\newblock {SWE}-bench: Can language models resolve real-world github issues?
\newblock In \emph{The Twelfth International Conference on Learning Representations}, 2024.
\newblock URL \url{https://openreview.net/forum?id=VTF8yNQM66}.

\bibitem[Kadavath et~al.(2022)Kadavath, Conerly, Askell, Henighan, Drain, Perez, Schiefer, Hatfield-Dodds, DasSarma, Tran-Johnson, et~al.]{kadavath2022language}
Saurav Kadavath, Tom Conerly, Amanda Askell, Tom Henighan, Dawn Drain, Ethan Perez, Nicholas Schiefer, Zac Hatfield-Dodds, Nova DasSarma, Eli Tran-Johnson, et~al.
\newblock Language models (mostly) know what they know.
\newblock \emph{arXiv preprint arXiv:2207.05221}, 2022.

\bibitem[{Kinde}(2024)]{kinde2024pricing}
{Kinde}.
\newblock Ai token pricing optimization: Dynamic cost management for llm-powered saas, 2024.
\newblock URL \url{https://kinde.com/learn/billing/billing-for-ai/ai-token-pricing-optimization-dynamic-cost-management-for-llm-powered-saas}.

\bibitem[Liu et~al.(2025)Liu, Wang, Miao, Hsu, Yan, Chen, Han, Xu, Chen, Jiang, et~al.]{liu2025budget}
Tengxiao Liu, Zifeng Wang, Jin Miao, I~Hsu, Jun Yan, Jiefeng Chen, Rujun Han, Fangyuan Xu, Yanfei Chen, Ke~Jiang, et~al.
\newblock Budget-aware tool-use enables effective agent scaling.
\newblock \emph{arXiv preprint arXiv:2511.17006}, 2025.

\bibitem[Liu et~al.(2023{\natexlab{a}})Liu, Xu, and McAuley]{liu2023repobenchbenchmarkingrepositorylevelcode}
Tianyang Liu, Canwen Xu, and Julian McAuley.
\newblock Repobench: Benchmarking repository-level code auto-completion systems, 2023{\natexlab{a}}.
\newblock URL \url{https://arxiv.org/abs/2306.03091}.

\bibitem[Liu et~al.(2023{\natexlab{b}})Liu, Yu, Zhang, Xu, Lei, Lai, Gu, Ding, Men, Yang, Zhang, Deng, Zeng, Du, Zhang, Shen, Zhang, Su, Sun, Huang, Dong, and Tang]{liu2023agentbenchevaluatingllmsagents}
Xiao Liu, Hao Yu, Hanchen Zhang, Yifan Xu, Xuanyu Lei, Hanyu Lai, Yu~Gu, Hangliang Ding, Kaiwen Men, Kejuan Yang, Shudan Zhang, Xiang Deng, Aohan Zeng, Zhengxiao Du, Chenhui Zhang, Sheng Shen, Tianjun Zhang, Yu~Su, Huan Sun, Minlie Huang, Yuxiao Dong, and Jie Tang.
\newblock Agentbench: Evaluating llms as agents, 2023{\natexlab{b}}.
\newblock URL \url{https://arxiv.org/abs/2308.03688}.

\bibitem[Madaan et~al.(2023)Madaan, Tandon, Gupta, Hallinan, Gao, Wiegreffe, Alon, Dziri, Prabhumoye, Yang, et~al.]{madaan2023self}
Aman Madaan, Niket Tandon, Prakhar Gupta, Skyler Hallinan, Luyu Gao, Sarah Wiegreffe, Uri Alon, Nouha Dziri, Shrimai Prabhumoye, Yiming Yang, et~al.
\newblock Self-refine: Iterative refinement with self-feedback.
\newblock \emph{Advances in neural information processing systems}, 36:\penalty0 46534--46594, 2023.

\bibitem[{OpenAI}(2025)]{openai2025codex}
{OpenAI}.
\newblock Introducing codex, 2025.
\newblock URL \url{https://openai.com/blog/introducing-codex}.

\bibitem[Salim et~al.(2026)Salim, Latendresse, Khatoonabadi, and Shihab]{salim2026tokenomics}
Mohamad Salim, Jasmine Latendresse, SayedHassan Khatoonabadi, and Emad Shihab.
\newblock Tokenomics: Quantifying where tokens are used in agentic software engineering.
\newblock \emph{arXiv preprint arXiv:2601.14470}, 2026.

\bibitem[Snell et~al.(2024)Snell, Lee, Xu, and Kumar]{snell2024scalingllmtesttimecompute}
Charlie Snell, Jaehoon Lee, Kelvin Xu, and Aviral Kumar.
\newblock Scaling llm test-time compute optimally can be more effective than scaling model parameters, 2024.
\newblock URL \url{https://arxiv.org/abs/2408.03314}.

\bibitem[Wang et~al.(2023)Wang, Xie, Jiang, Mandlekar, Xiao, Zhu, Fan, and Anandkumar]{wang2023voyager}
Guanzhi Wang, Yuqi Xie, Yunfan Jiang, Ajay Mandlekar, Chaowei Xiao, Yuke Zhu, Linxi Fan, and Anima Anandkumar.
\newblock Voyager: An open-ended embodied agent with large language models.
\newblock \emph{arXiv preprint arXiv:2305.16291}, 2023.

\bibitem[Wang et~al.(2025{\natexlab{a}})Wang, Hu, Liu, Zhu, Hou, Huang, Zhang, Yang, Liu, Zhang, et~al.]{wang2025efficient}
Ningning Wang, Xavier Hu, Pai Liu, He~Zhu, Yue Hou, Heyuan Huang, Shengyu Zhang, Jian Yang, Jiaheng Liu, Ge~Zhang, et~al.
\newblock Efficient agents: Building effective agents while reducing cost.
\newblock \emph{arXiv preprint arXiv:2508.02694}, 2025{\natexlab{a}}.

\bibitem[Wang et~al.(2025{\natexlab{b}})Wang, Tang, Jiang, Chen, Wang, and He]{wang2025agenttaxo}
Qian Wang, Zhenheng Tang, Zichen Jiang, Nuo Chen, Tianyu Wang, and Bingsheng He.
\newblock Agenttaxo: Dissecting and benchmarking token distribution of llm multi-agent systems.
\newblock In \emph{ICLR 2025 Workshop on Foundation Models in the Wild}, 2025{\natexlab{b}}.

\bibitem[Wang et~al.(2024)Wang, Chen, Yuan, Zhang, Li, Peng, and Ji]{wang2024executable}
Xingyao Wang, Yangyi Chen, Lifan Yuan, Yizhe Zhang, Yunzhu Li, Hao Peng, and Heng Ji.
\newblock Executable code actions elicit better llm agents.
\newblock In \emph{Forty-first International Conference on Machine Learning}, 2024.

\bibitem[Wang et~al.(2025{\natexlab{c}})Wang, Li, Song, Xu, Tang, Zhuge, Pan, Song, Li, Singh, Tran, Li, Ma, Zheng, Qian, Shao, Muennighoff, Zhang, Hui, Lin, Brennan, Peng, Ji, and Neubig]{wang2025openhands}
Xingyao Wang, Boxuan Li, Yufan Song, Frank~F. Xu, Xiangru Tang, Mingchen Zhuge, Jiayi Pan, Yueqi Song, Bowen Li, Jaskirat Singh, Hoang~H. Tran, Fuqiang Li, Ren Ma, Mingzhang Zheng, Bill Qian, Yanjun Shao, Niklas Muennighoff, Yizhe Zhang, Binyuan Hui, Junyang Lin, Robert Brennan, Hao Peng, Heng Ji, and Graham Neubig.
\newblock Openhands: An open platform for {AI} software developers as generalist agents.
\newblock In \emph{The Thirteenth International Conference on Learning Representations}, 2025{\natexlab{c}}.
\newblock URL \url{https://openreview.net/forum?id=OJd3ayDDoF}.

\bibitem[Wu et~al.(2025)Wu, Sun, Li, Welleck, and Yang]{wu2025inferencescalinglawsempirical}
Yangzhen Wu, Zhiqing Sun, Shanda Li, Sean Welleck, and Yiming Yang.
\newblock Inference scaling laws: An empirical analysis of compute-optimal inference for problem-solving with language models, 2025.
\newblock URL \url{https://arxiv.org/abs/2408.00724}.

\bibitem[Yang et~al.(2025)Yang, Ma, Lin, and Wei]{yang2025towards}
Wenkai Yang, Shuming Ma, Yankai Lin, and Furu Wei.
\newblock Towards thinking-optimal scaling of test-time compute for {LLM} reasoning.
\newblock In \emph{The Thirty-ninth Annual Conference on Neural Information Processing Systems}, 2025.
\newblock URL \url{https://openreview.net/forum?id=6ICFqmixlS}.

\bibitem[Zeng et~al.(2025)Zeng, Cheng, Yin, Zhou, and Qiu]{zeng2025revisiting}
Zhiyuan Zeng, Qinyuan Cheng, Zhangyue Yin, Yunhua Zhou, and Xipeng Qiu.
\newblock Revisiting the test-time scaling of o1-like models: Do they truly possess test-time scaling capabilities?
\newblock In \emph{Proceedings of the 63rd Annual Meeting of the Association for Computational Linguistics (Volume 1: Long Papers)}, pp.\  4651--4665, 2025.

\end{thebibliography}
\bibliographystyle{colm2026_conference}
\newpage
\appendix

\section{Output Token Analyses}
This appendix presents complementary analyses using \emph{output tokens} in place of input tokens.
Across all settings, the output-token results closely mirror the trends reported in the main text: accuracy decreases as output-token cost increases, and higher-cost runs are associated with a sharp rise in repeated file \emph{view} and \emph{modify} actions.
These findings reinforce the conclusion that excessive computation is primarily driven by redundant agent behavior rather than productive progress, and that the inverse accuracy--cost relationship is not specific to input tokens.

\label{sec:completion-tokens}

\begin{figure}[H]
  \centering
  \begin{subfigure}{0.47\textwidth}
    \centering
    \includegraphics[width=\linewidth]{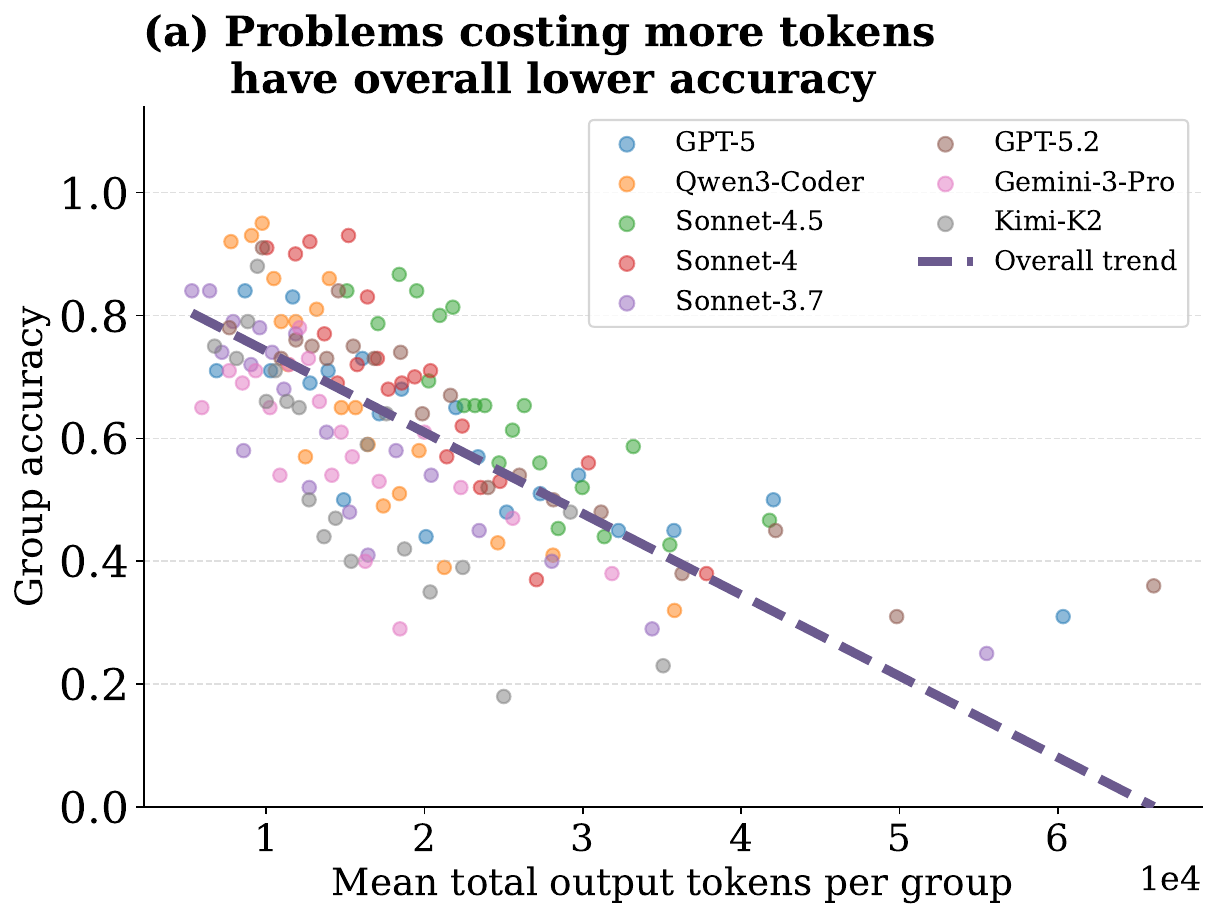}
    \subcaption{Higher output-token usage is associated with lower task accuracy.}
    \label{fig:completion-scatter-app}
  \end{subfigure}
  \hfill
  \begin{subfigure}{0.47\textwidth}
    \centering
    \includegraphics[width=\linewidth]{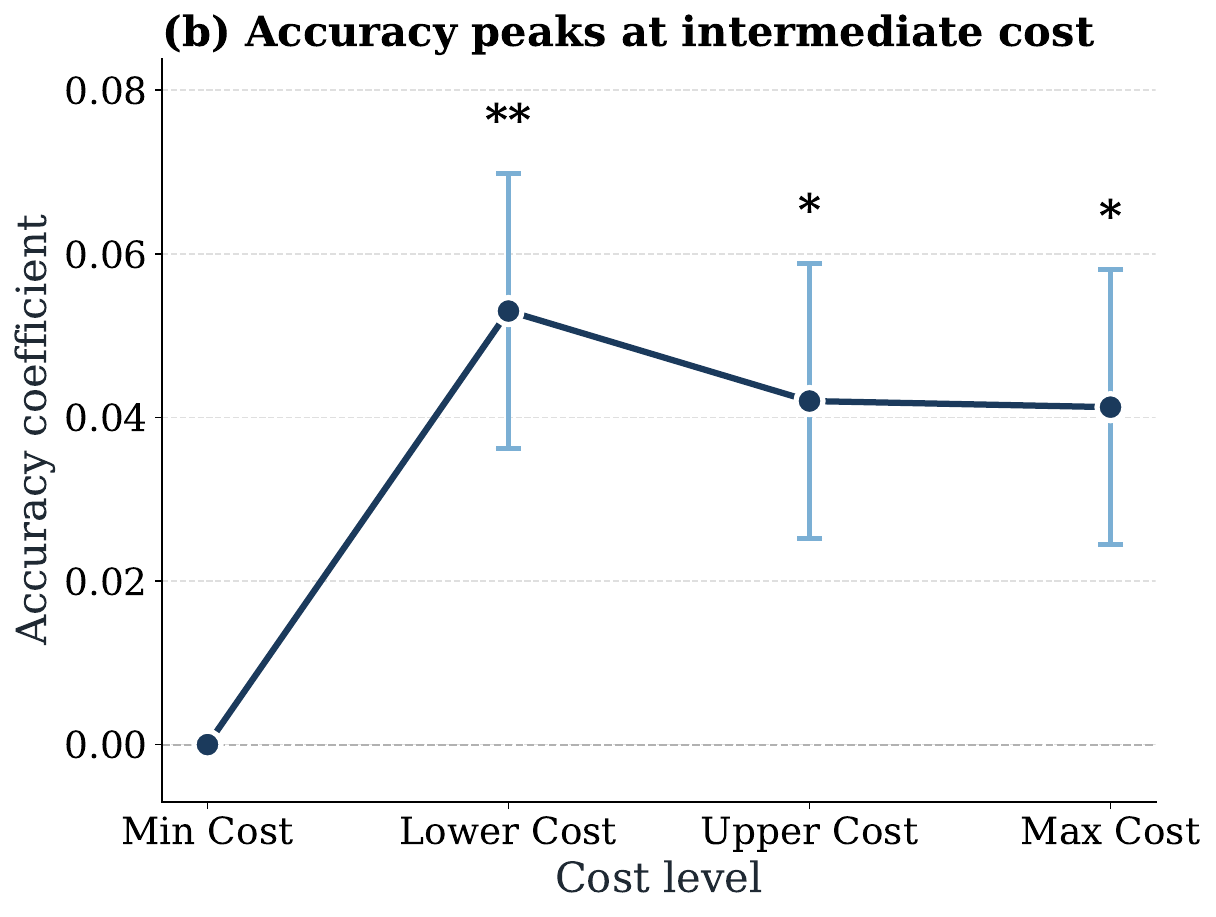}
    \subcaption{Accuracy improves at intermediate cost levels but declines at higher costs.}
    \label{fig:completion-regression-app}
  \end{subfigure}

  \vspace{0.6em}

  \begin{subfigure}{0.47\textwidth}
    \centering
    \includegraphics[width=\linewidth]{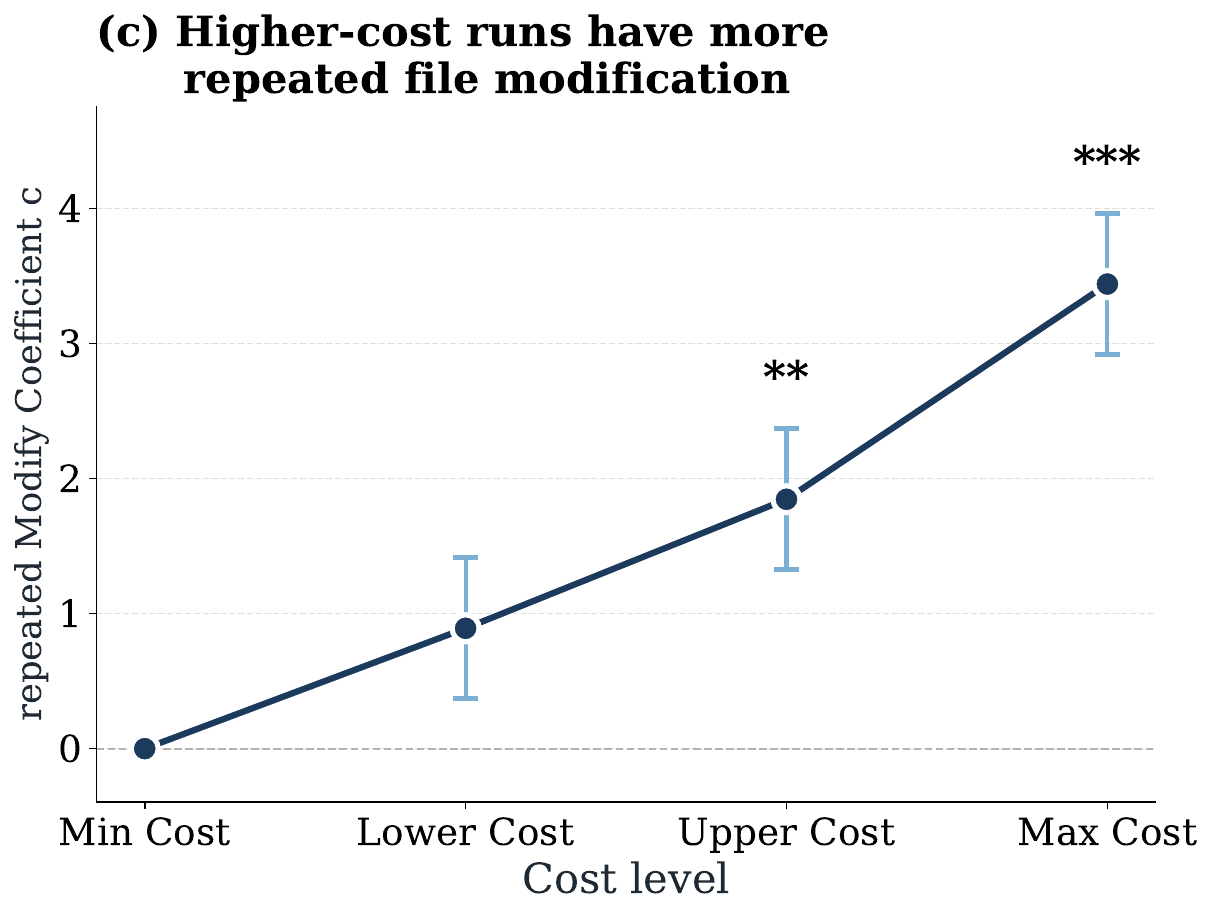}
    \subcaption{Repeated file \textit{modification} actions increase with task cost.}
    \label{fig:completion-repeated-modify-app}
  \end{subfigure}
  \hfill
  \begin{subfigure}{0.47\textwidth}
    \centering
    \includegraphics[width=\linewidth]{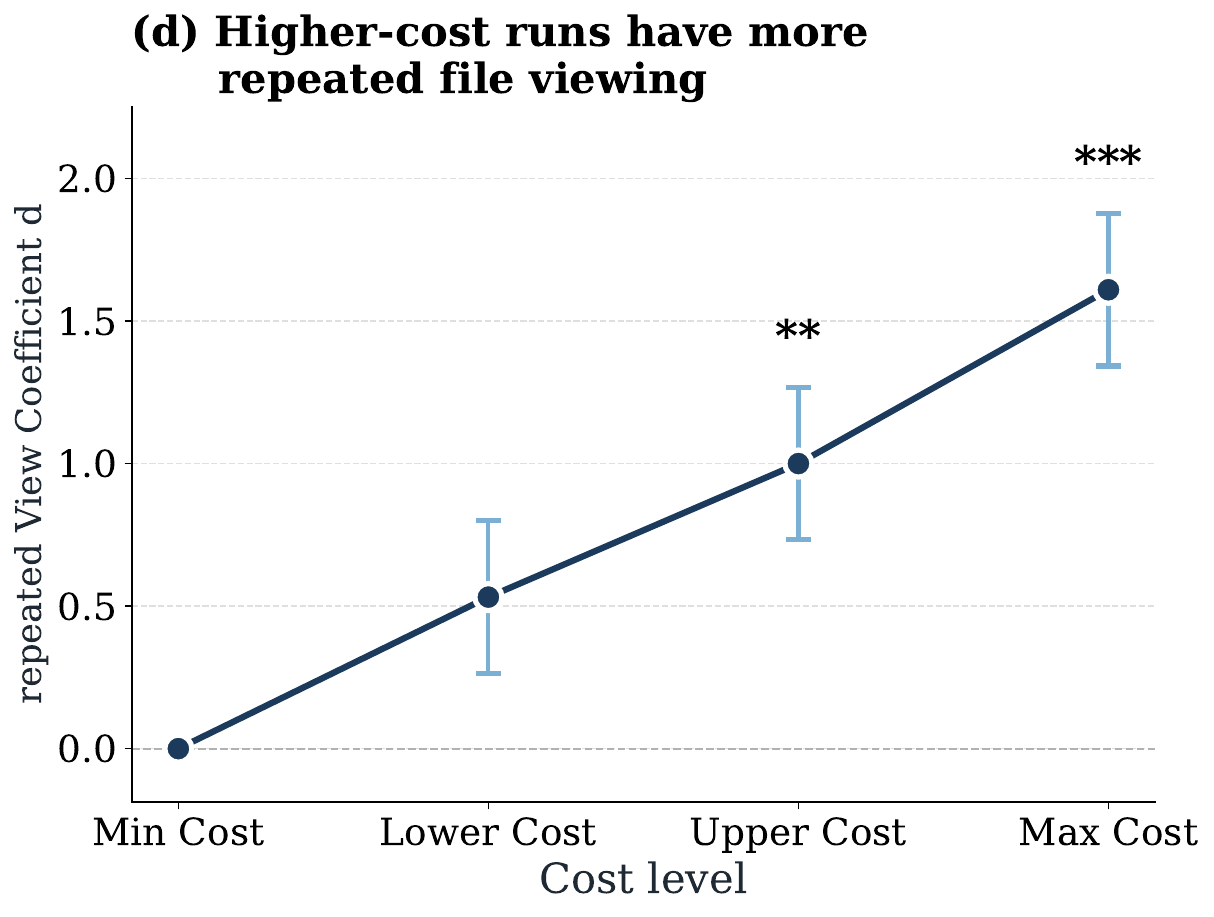}
    \subcaption{Repeated file \textit{viewing} actions increase with task cost.}
    \label{fig:completion-repeated-view-app}
  \end{subfigure}

  \caption{
  \textbf{Output-token analyses.}
    Top: Group-level accuracy vs. mean output tokens by model, with mixed-effects trends controlling for model identity.Bottom: Mixed-effects coefficients (vs. MinCost) for repeated modify and view actions across cost levels.Higher costs reduce accuracy and sharply increase redundant file operations.
  }
  \label{fig:completion-token-appendix}
\end{figure}

\section{Cost Calculation Details}
\label{app:cost-formulas}

\subsection{Explicit Caching Models (Claude Models)}
\label{app:cost-claude}
\begin{equation}
\label{eq:prompt-nc}
\text{Input}_{\text{non-cached}} 
= \text{Input}_{\text{total}} - \text{CacheRead}.
\end{equation}

\begin{align}
\label{eq:cost-round}
\text{Cost}_{\text{round}} =\;&
(\text{Input}_{\text{non-cached}} \times r_{\text{in}}) \notag \\
&+ (\text{Output} \times r_{\text{out}}) \notag \\
&+ (\text{CacheCreation} \times r_{\text{cache\_create}}) \notag \\
&+ (\text{CacheRead} \times r_{\text{cache\_read}}).
\end{align}

where $r_{\text{in}}$ is the base input rate, $r_{\text{out}}$ is the output rate, $r_{\text{cache\_create}}$ is the cache creation rate (5-minute writes in our setting), and $r_{\text{cache\_read}}$ is the cache read rate.

\subsection{Implicit Cache (GPT5 and alike)}
\label{app:cost-gpt5}

For GPT-5 models, we use OpenAI’s implicit caching mechanism. The API reports
cached input tokens automatically, without explicit cache creation. At the
time of our experiments, the official pricing is:
\emph{Input: \$1.250 / 1M tokens}, \emph{Cached input: \$0.125 / 1M tokens},
and \emph{Output: \$10.000 / 1M tokens}.\footnote{See the official pricing documentation:
\url{https://openai.com/index/introducing-gpt-5/}}

The non-cached prompt is
\begin{equation}
\text{Input}_{\text{non-cached}}
= \text{Input}_{\text{total}}
- \text{CacheRead}_{\text{implicit}}.
\end{equation}

The total cost is
\begin{align}
\text{Cost}_{\text{round}} =\;&
(\text{Input}_{\text{non-cached}} \times r_{\text{in}}) \notag \\
&+ (\text{CacheRead}_{\text{implicit}} \times 0.2\, r_{\text{in}}) \notag \\
&+ (\text{Output} \times r_{\text{out}}).
\end{align}

\section{Prompt for Self-Prediction by the Same Agent}\label{sec:agent_setting}

\subsection{System Prompt}
\label{prompt:agent-zero-shot-prompt}

\begin{tcolorbox}[
    colback=yellow!5,
    colframe=yellow!60!black,
    breakable,
    enhanced,
    left=0pt, right=0pt, top=2pt, bottom=2pt,
    before upper=\setlength{\parskip}{6pt} 
] 

\paragraph{System Role}

You are an \textbf{OpenHands agent}, a helpful AI assistant that can interact with a computer to estimate token costs for software engineering tasks.

\paragraph{Role Definition}

Your primary role in this configuration is \textbf{token estimation} for SWE-bench--style software engineering tasks.
You should thoroughly understand the task and repository using normal tools (e.g., bash, IPython, browsing, file viewing) as needed for analysis.
However, instead of implementing a fix, you must \textbf{output a final token cost estimate} for completing the task with this agent.
If the user asks a question (e.g., ``why is X happening''), answer directly.

\paragraph{Token Estimation Objective}

Your goal is to predict how many LLM tokens will be consumed in total if this agent were to complete the task end-to-end.
You must provide:
\begin{itemize}
  \item \texttt{predicted\_input\_tokens}: Estimated total tokens sent to the LLM
  \item \texttt{predicted\_output\_tokens}: Estimated total tokens generated by the LLM
  \item \texttt{predicted\_total\_tokens}: Sum of input and output tokens
  \item \texttt{confidence}: A number in $[0,1]$
  \item \texttt{breakdown\_by\_phase}: A JSON object breaking down input/output tokens across phases (e.g., repo\_cloning, initial\_reading, test\_setup, debugging, coding\_iterations, verification, review\_cleanup)
\end{itemize}

\paragraph{Critical Constraint: Token Estimation Only}

\textbf{IMPORTANT:}
\begin{itemize}
  \item You are a \textbf{token estimation agent}, not a problem-solving agent.
  \item Your only goal is to estimate token costs, not to fix bugs or implement features.
  \item You must call the \texttt{finish} tool with a JSON estimate.
  \item Never output actual code changes.
  \item If you find yourself writing code to fix the problem, stop and call \texttt{finish}.
\end{itemize}

\textbf{Examples of what you should do:}
\begin{itemize}
  \item Explore the codebase to understand the problem.
  \item Run tests to observe current failures.
  \item Analyze code complexity and required changes.
  \item Estimate token usage for the complete solution.
  \item Call \texttt{finish} with a JSON estimate.
\end{itemize}

\textbf{Examples of what you must not do:}
\begin{itemize}
  \item Write actual code fixes.
  \item Modify source files.
  \item Create test files.
  \item Commit changes.
  \item Call \texttt{finish} with code patches.
\end{itemize}

\paragraph{Efficiency Guidelines}

\begin{itemize}
  \item Each action you take is somewhat expensive.
  \item Wherever possible, combine multiple actions into a single command.
  \item Use efficient tools such as \texttt{find}, \texttt{grep}, and \texttt{git} with appropriate filters to minimize unnecessary operations.
\end{itemize}

\paragraph{File System Guidelines}

\begin{itemize}
  \item Do not assume user-provided paths are relative to the current working directory.
  \item Use efficient tools (e.g., \texttt{grep}, \texttt{find}, \texttt{cat}) to inspect files.
  \item You may read files for analysis, but must not modify source files.
  \item Delete any temporary files created strictly for exploration or testing.
\end{itemize}

\paragraph{Estimation Workflow}

\begin{enumerate}
  \item \textbf{Exploration:} Explore relevant files and understand the task context.
  \item \textbf{Analysis:} Consider multiple solution approaches and estimate token costs for each phase.
  \item \textbf{Testing (for estimation purposes):}
  \begin{itemize}
    \item You may run existing tests to understand failures and complexity.
    \item Do not create new tests or modify existing tests.
    \item If tests cannot be run, estimate based on code analysis.
  \end{itemize}
  \item \textbf{Estimation:}
  \begin{itemize}
    \item Synthesize findings into a token usage plan.
    \item Base estimates on repository size, test runs, iterations, and tool usage.
    \item Consider all phases: repo cloning, initial reading, test setup, debugging, coding iterations, verification, and review cleanup.
    \item Delete any temporary exploration files.
  \end{itemize}
  \item \textbf{Output:} Produce the estimate as a JSON object and call \texttt{finish}.
\end{enumerate}

\paragraph{Security}

\begin{itemize}
  \item Only use \texttt{GITHUB\_TOKEN} or other credentials when explicitly requested and expected.
  \item Use APIs for GitHub or other platforms unless browsing is required.
\end{itemize}

\paragraph{Environment Setup}

\begin{itemize}
  \item If an application is missing, install it and rerun the command.
  \item If dependencies are missing:
  \begin{enumerate}
    \item Look for dependency files (e.g., \texttt{requirements.txt}, \texttt{pyproject.toml}, \texttt{package.json}).
    \item Install dependencies using those files if present.
    \item Install individual packages only if no dependency files exist.
  \end{enumerate}
\end{itemize}

\paragraph{Troubleshooting}

\begin{itemize}
  \item If issues arise during exploration or testing:
  \begin{enumerate}
    \item Document the issue in your reasoning.
    \item Factor added complexity into token estimates.
    \item Continue estimation with available information.
  \end{enumerate}
\end{itemize}

\paragraph{Token Estimation Output Format}

\begin{verbatim}
{
  "predicted_input_tokens": <integer>,
  "predicted_output_tokens": <integer>,
  "predicted_total_tokens": <integer>,
  "confidence": <float between 0 and 1>,
  "breakdown_by_phase": {
    "repo_cloning": { ... },
    "initial_reading": { ... },
    "test_setup": { ... },
    "debugging": { ... },
    "coding_iterations": { ... },
    "verification": { ... },
    "review_cleanup": { ... }
  }
}
\end{verbatim}

\end{tcolorbox}

\newpage

\subsection{In-context Example and User Instruction}
\label{prompt:agent-zero-shot-grad-prompt}

\begin{tcolorbox}[
    colback=orange!5, 
    colframe=orange!60!black, 
    breakable,
    enhanced,
    left=0pt, right=0pt, top=2pt, bottom=2pt,
    before upper=\setlength{\parskip}{6pt} 
] 

\paragraph{Running Example (Token Cost Estimation for a Software Engineering Task)}

\noindent\rule{\linewidth}{0.4pt}

\noindent\textbf{START OF EXAMPLE}

\begin{quote}
\textbf{USER:} I've uploaded a python code repository in the directory \texttt{example\_repo}. You are a TOKEN ESTIMATION agent. Estimate the token cost to fix the following issue description:

\medskip
\textbf{\textless issue\_description\textgreater}\\
Fix a bug where the version comparison function fails for development versions (e.g., \texttt{"1.2.3.dev0"}) when used across multiple modules in a large codebase with complex dependencies.\\
\textbf{\textless/issue\_description\textgreater}

\medskip
\textbf{ASSISTANT:} I'll explore the codebase to understand the problem and estimate token costs.

\medskip
\textbf{\textless function=execute\_bash\textgreater}\\
\textbf{\textless parameter=command\textgreater}\\
\texttt{cd example\_repo \&\& find . -name "*.py" -type f | head -50}\\
\textbf{\textless/parameter\textgreater}\\
\textbf{\textless/function\textgreater}

\medskip
\textbf{USER:} EXECUTION RESULT of [execute\_bash]:
\begin{verbatim}
./src/core/version.py
./src/core/utils.py
./src/api/endpoints.py
./src/api/middleware.py
./src/processing/transformers.py
./src/processing/validators.py
./src/integration/external_api.py
./src/integration/adapters.py
./tests/unit/test_version.py
./tests/unit/test_integration.py
./tests/integration/test_api.py
./tests/functional/test_workflows.py
... (45 more files)
\end{verbatim}

\textbf{ASSISTANT:} This is a larger codebase. Let me explore the structure and find where version comparison is used:

\medskip
\textbf{\textless function=execute\_bash\textgreater}\\
\textbf{\textless parameter=command\textgreater}\\
\texttt{cd example\_repo \&\& grep -r "compare\_versions\textbackslash|LooseVersion" --include="*.py" | head -30}\\
\textbf{\textless/parameter\textgreater}\\
\textbf{\textless/function\textgreater}

\medskip
\textbf{USER:} EXECUTION RESULT of [execute\_bash]:
\begin{Verbatim}[breaklines, breakanywhere]
./src/core/version.py:from distutils.version import LooseVersion
./src/core/version.py:def compare_versions(v1, v2):
./src/api/endpoints.py:from src.core.version import compare_versions
./src/processing/transformers.py:from src.core.version import compare_versions
./src/integration/external_api.py:from src.core.version import compare_versions
./src/integration/adapters.py:from src.core.version import compare_versions
... (12 more usages)
\end{Verbatim}

\textbf{ASSISTANT:} The version comparison function is used in many places. Let me read the main version module and understand the dependencies:

\medskip
\textbf{\textless function=execute\_bash\textgreater}\\
\textbf{\textless parameter=command\textgreater}\\
\texttt{cat example\_repo/src/core/version.py}\\
\textbf{\textless/parameter\textgreater}\\
\textbf{\textless/function\textgreater}

\medskip
\textbf{USER:} EXECUTION RESULT of [execute\_bash]:
\begin{verbatim}
from distutils.version import LooseVersion

def compare_versions(v1, v2):
    return LooseVersion(v1) >= LooseVersion(v2)
\end{verbatim}

\textbf{ASSISTANT:} Now let me check how it's used in the integration modules to understand the impact:

\medskip
\textbf{\textless function=execute\_bash\textgreater}\\
\textbf{\textless parameter=command\textgreater}\\
\texttt{cat example\_repo/src/integration/external\_api.py | head -100}\\
\textbf{\textless/parameter\textgreater}\\
\textbf{\textless/function\textgreater}

\medskip
\textbf{USER:} EXECUTION RESULT of [execute\_bash]:
\begin{Verbatim}[breaklines, breakanywhere]
import requests
from src.core.version import compare_versions
# ... (large file with 500+ lines using compare_versions in multiple places)
\end{Verbatim}

\textbf{ASSISTANT:} Let me check the test files to understand the expected behavior and test coverage:

\medskip
\textbf{\textless function=execute\_bash\textgreater}\\
\textbf{\textless parameter=command\textgreater}\\
\texttt{cat example\_repo/tests/unit/test\_version.py}\\
\textbf{\textless/parameter\textgreater}\\
\textbf{\textless/function\textgreater}

\medskip
\textbf{USER:} EXECUTION RESULT of [execute\_bash]:
\begin{Verbatim}[breaklines, breakanywhere]
import pytest
from src.core.version import compare_versions

def test_compare_versions():
    assert compare_versions("1.2.3", "1.2.2") == True
    assert compare_versions("1.2.3.dev0", "1.2.2") == True  # This currently fails

def test_compare_versions_edge_cases():
    # ... (multiple edge case tests)
\end{Verbatim}

\textbf{ASSISTANT:} Let me run the existing tests to see the current failures:

\medskip
\textbf{\textless function=execute\_bash\textgreater}\\
\textbf{\textless parameter=command\textgreater}\\
\texttt{cd example\_repo \&\& python -m pytest tests/unit/test\_version.py -v}\\
\textbf{\textless/parameter\textgreater}\\
\textbf{\textless/function\textgreater}

\medskip
\textbf{USER:} EXECUTION RESULT of [execute\_bash]:
\begin{Verbatim}[breaklines, breakanywhere]
FAILED tests/unit/test_version.py::test_compare_versions - ValueError: invalid version number '1.2.3.dev0'
FAILED tests/unit/test_version.py::test_compare_versions_edge_cases - ValueError: invalid version number '1.2.3.dev0'
\end{Verbatim}

\textbf{ASSISTANT:} Let me also check integration tests that might be affected:

\medskip
\textbf{\textless function=execute\_bash\textgreater}\\
\textbf{\textless parameter=command\textgreater}\\
\texttt{cd example\_repo \&\& python -m pytest tests/integration/ -v --tb=short 2>\&1 | head -50}\\
\textbf{\textless/parameter\textgreater}\\
\textbf{\textless/function\textgreater}

\medskip
\textbf{USER:} EXECUTION RESULT of [execute\_bash]:
\begin{Verbatim}[breaklines, breakanywhere]
FAILED tests/integration/test_api.py::test_version_checking - ValueError: invalid version number '1.2.3.dev0'
FAILED tests/integration/test_api.py::test_external_api_integration - ValueError: invalid version number '1.2.3.dev0'
... (8 more failures)
\end{Verbatim}

\textbf{ASSISTANT:} Based on my exploration, I can now estimate the token costs. This is a complex issue that requires:
\begin{enumerate}
  \item Understanding the problem across a large codebase with 50+ files
  \item Reading and analyzing multiple files that use the version comparison function
  \item Finding an alternative solution that maintains backward compatibility
  \item Updating all affected modules and ensuring no regressions
  \item Running comprehensive test suites (unit, integration, functional)
  \item Multiple debugging iterations to handle edge cases
  \item Verifying the fix works across all integration points
\end{enumerate}

\medskip
\textbf{\textless function=finish\textgreater}\\
\textbf{\textless parameter=message\textgreater}
\begin{Verbatim}[breaklines, breakanywhere]
{"predicted_input_tokens": 1250000, "predicted_output_tokens": 18500, "predicted_total_tokens": 1268500, "confidence": 0.82, "breakdown_by_phase": {"repo_cloning": {"input_tokens": 5000, "output_tokens": 800, "total_tokens": 5800, "description": "Exploring large codebase structure, finding all Python files, understanding directory layout"}, "initial_reading": {"input_tokens": 450000, "output_tokens": 3500, "total_tokens": 453500, "description": "Reading issue description, searching for version comparison usage across codebase, reading version.py (~2000 tokens), reading 8+ files that import it (~15000 tokens each), reading test files (~8000 tokens), understanding dependencies and integration points"}, "test_setup": {"input_tokens": 120000, "output_tokens": 2500, "total_tokens": 122500, "description": "Running unit tests, integration tests, functional tests, analyzing failure patterns, understanding test coverage requirements"}, "debugging": {"input_tokens": 280000, "output_tokens": 4200, "total_tokens": 284200, "description": "Analyzing solution options (packaging.version vs custom parser), checking library availability, understanding backward compatibility requirements, evaluating impact on all integration points"}, "coding_iterations": {"input_tokens": 320000, "output_tokens": 4800, "total_tokens": 324800, "description": "Implementing fix in version.py, updating all affected modules, handling edge cases, multiple test-run-fix cycles, ensuring no regressions in existing functionality"}, "verification": {"input_tokens": 65000, "output_tokens": 2200, "total_tokens": 67200, "description": "Running full test suite (unit + integration + functional), verifying edge cases, checking integration points, validating backward compatibility"}, "review_cleanup": {"input_tokens": 5000, "output_tokens": 500, "total_tokens": 5500, "description": "Final code review, documentation updates, ensuring code quality standards"}}}
\end{Verbatim}
\textbf{\textless/parameter\textgreater}\\
\textbf{\textless/function\textgreater}
\end{quote}

\noindent\textbf{END OF EXAMPLE}

\noindent\rule{\linewidth}{0.4pt}

\paragraph{New Task Description}

Do \textbf{not} assume the environment is the same as in the example above.

\medskip
\textbf{Uploaded files:}
\begin{verbatim}
/workspace/{{ workspace_dir_name }}
\end{verbatim}

\noindent I've uploaded a python code repository in the directory \texttt{\{\{ workspace\_dir\_name \}\}}. You are a TOKEN ESTIMATION agent. Estimate the token cost to fix the following issue description:

\medskip
\textbf{\textless issue\_description\textgreater}\\
\texttt{\{\{ instance.problem\_statement \}\}}\\
\textbf{\textless/issue\_description\textgreater}

\paragraph{Important Constraints}

\begin{itemize}
  \item You are a TOKEN ESTIMATION agent, NOT a problem-solving agent.
  \item Your ONLY goal is to estimate token costs, NOT to fix bugs or implement features.
  \item You MUST call the finish tool with a JSON estimate, NEVER with actual code changes.
\end{itemize}

\paragraph{Granularity Rules}

\begin{itemize}
  \item Unit: \texttt{token} (integer).
  \item Avoid coarse rounding (e.g., 100s/1000s). Use a step size of 1.
\end{itemize}

\paragraph{Phases for Estimation}

Your task is to estimate how many LLM tokens would be consumed to solve this problem if a coding agent were to complete it end-to-end.

\medskip
\textbf{Phase 1. Exploration:} Explore the codebase to understand the problem
\begin{enumerate}
  \item Read the problem description and understand what needs to be fixed
  \item Explore relevant files and directories to understand the codebase structure
  \item Search for key functions, classes, or variables related to the issue
  \item Identify the root cause and complexity of the problem
\end{enumerate}

\textbf{Phase 2. Analysis:} Analyze the complexity and required changes
\begin{enumerate}
  \item Assess the scope of changes needed (number of files, lines of code)
  \item Consider the debugging iterations likely needed
  \item Evaluate the testing complexity and iterations
  \item Estimate the number of tool calls and reasoning steps
\end{enumerate}

\textbf{Phase 3. Token Estimation:} Calculate token usage for the complete solution
\begin{enumerate}
  \item Estimate input tokens for:
  \begin{itemize}
    \item Repository exploration and file reading
    \item Code analysis and debugging
    \item Implementation iterations
    \item Testing and verification
  \end{itemize}
  \item Estimate output tokens for:
  \begin{itemize}
    \item Reasoning and analysis responses
    \item Code generation and explanations
    \item Debugging responses
    \item Test results interpretation
  \end{itemize}
  \item Calculate total tokens and confidence level
\end{enumerate}

\textbf{Phase 4. Finish:} Provide final token estimate
\begin{enumerate}
  \item Call the finish tool with a JSON object containing:
  \begin{itemize}
    \item \texttt{predicted\_input\_tokens}
    \item \texttt{predicted\_output\_tokens}
    \item \texttt{predicted\_total\_tokens}
    \item \texttt{confidence} (0--1)
    \item \texttt{breakdown\_by\_phase}
  \end{itemize}
\end{enumerate}

\paragraph{Reminder}

Remember: You are estimating COSTS, not implementing SOLUTIONS.
Do not write actual code fixes or modify any files.
Your final deliverable is a JSON token estimate, not a working solution.

\medskip
\noindent\rule{\linewidth}{0.4pt}

\noindent\textbf{END OF NEW TASK DESCRIPTION}

\paragraph{Output Constraint (Format)}

Please follow the format strictly. Please emit one and only one function call per message.

\end{tcolorbox}


\clearpage
\section{Self-Prediction Without In-Context Example}
\label{sec:no-icl}

To examine whether the observed underestimation is induced by the in-context demonstration used in our main setup, we conducted additional runs without providing any example. 

\begin{figure}[H]
  \centering
  \begin{subfigure}{0.49\textwidth}
    \centering
    \includegraphics[width=\linewidth]{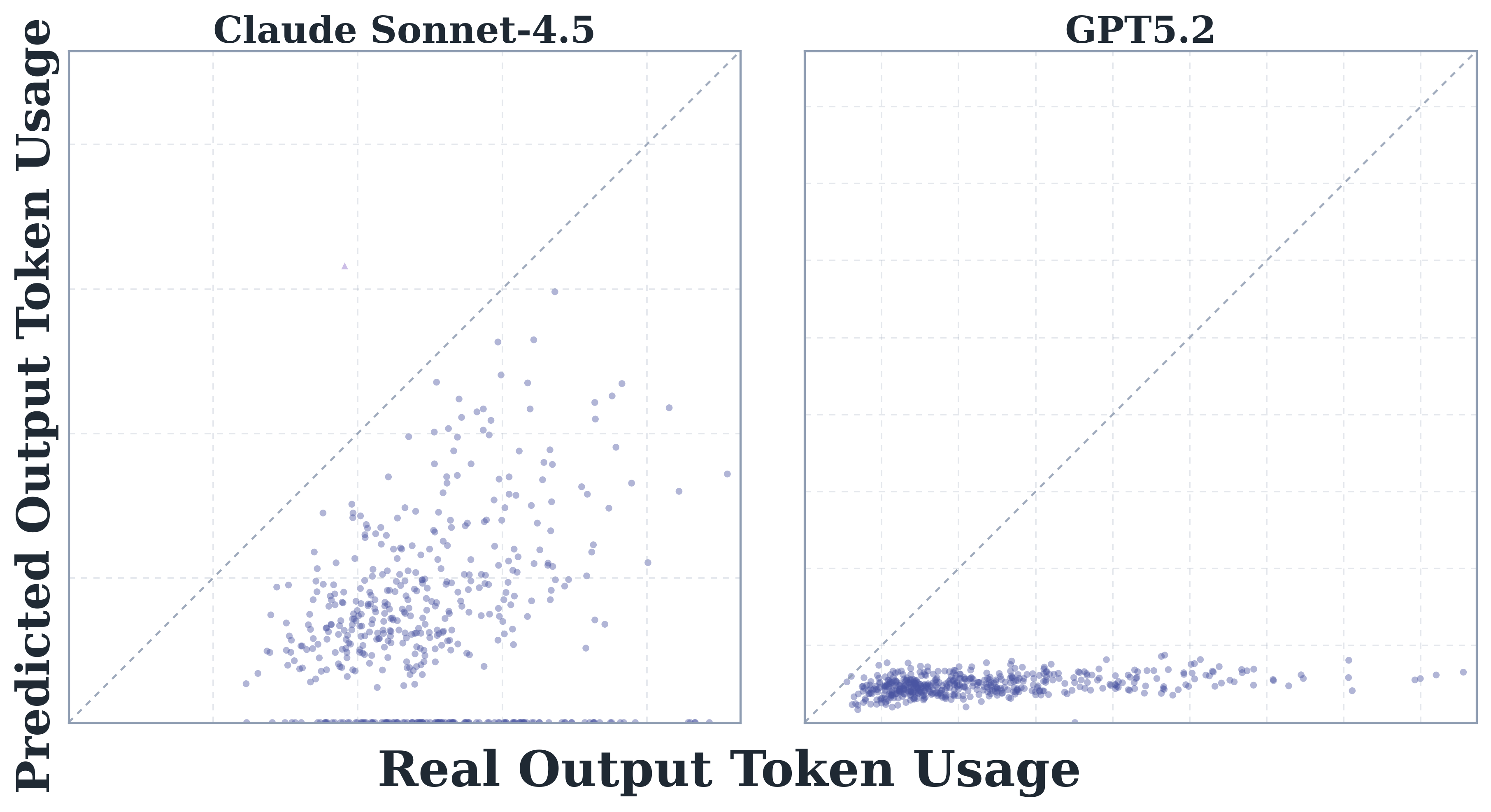}
    \caption{}
  \end{subfigure}
  \hfill
  \begin{subfigure}{0.49\textwidth}
    \centering
    \includegraphics[width=\linewidth]{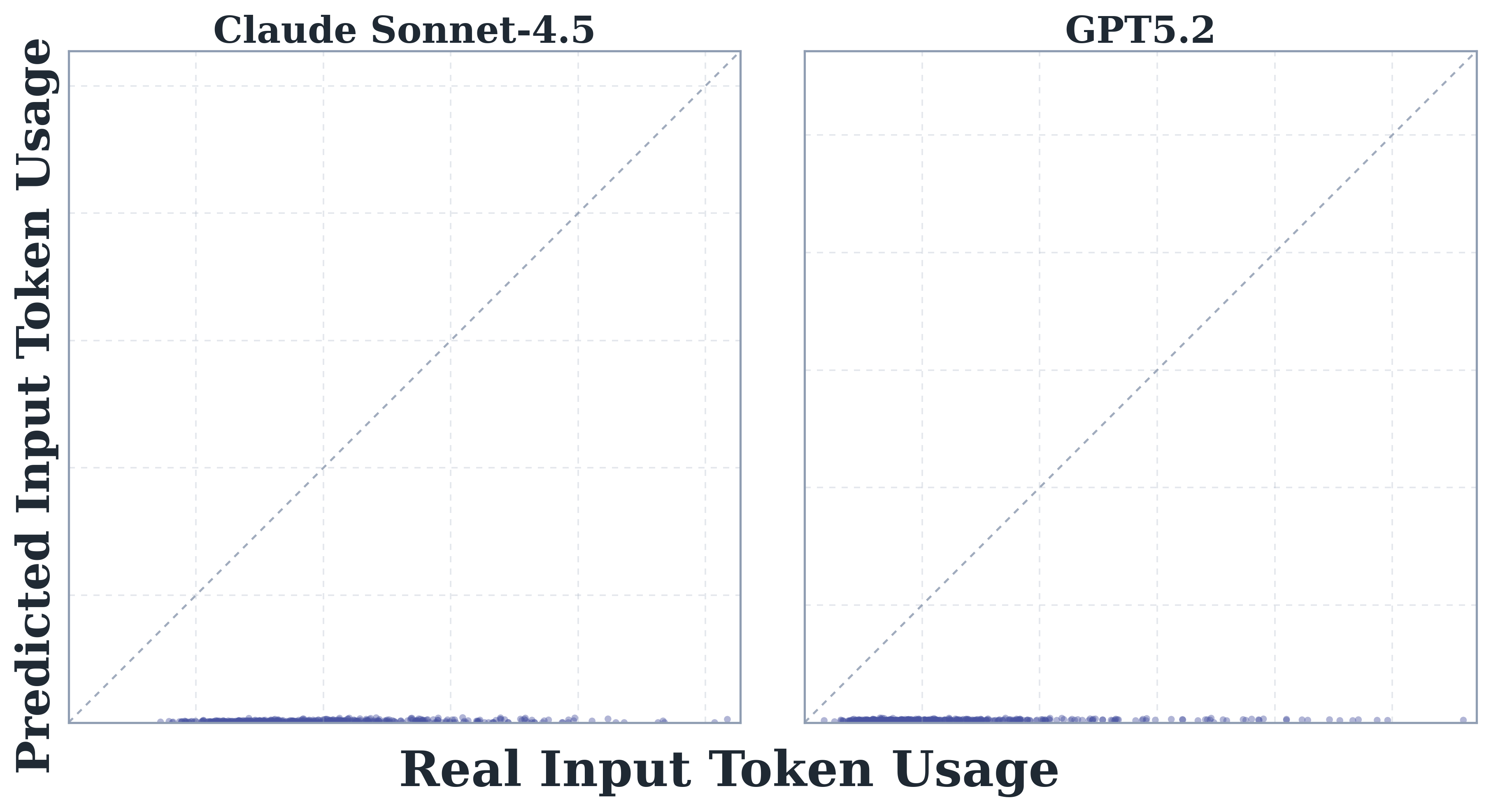}
    \caption{}
  \end{subfigure}
  \caption{\textbf{Self-prediction behavior without in-context demonstration (Sonnet~4.5 and GPT-5.2).}
  The dashed diagonal indicates perfect calibration. Underestimation remains pronounced.}
  \label{fig:no-icl-scatter}
\end{figure}

In practice, most models failed to consistently follow the instruction to perform token estimation without such demonstration; therefore, we report results for Sonnet~4.5 and GPT-5.2, which remained instruction-compliant. 
As shown in Figure~\ref{fig:no-icl-scatter}, underestimation persists, and becomes more severe, particularly for input tokens. 

Table~\ref{tab:no-icl-performance} further shows that correlation with real token usage degrades substantially without the in-context example. 
These results indicate that the downward bias is not caused by example-induced anchoring; instead, the demonstration improves calibration, while the underlying difficulty of anticipating long-horizon token growth remains.

\begin{table}[H]
\centering
\small
\resizebox{\linewidth}{!}{%
\begin{tabular}{l l c c c}
\toprule
\textbf{Model} & \textbf{Token} & \textbf{Corr w/ GT} &
\textbf{Corr (AbsErr, task cost)} &
\textbf{Corr (Pred cost, task cost)} \\
\midrule
Sonnet-4.5 & Input  & 0.1355 & 0.1155  & 0.1185 \\
Sonnet-4.5 & Output & 0.1229 & -0.4563 & 0.1185 \\
\midrule
GPT-5.2    & Input  & 0.1796 & 0.2243  & 0.2461 \\
GPT-5.2    & Output & 0.2130 & 0.0822  & 0.2461 \\
\bottomrule
\end{tabular}%
}
\caption{\textbf{Self-prediction performance without in-context example.}
Correlation metrics degrade compared to the main setup with demonstration.}
\label{tab:no-icl-performance}
\end{table}

\end{document}